\documentclass[10pt,twocolumn,letterpaper]{article}

\usepackage[table]{xcolor}
\usepackage{wacv}
\usepackage{times}
\usepackage{epsfig}
\usepackage{tikz}
\usepackage{graphicx}
\usepackage{amsmath}
\usepackage{amssymb}

\wacvfinalcopy 

\def\wacvPaperID{628} 

\usepackage[pagebackref=true,breaklinks=true,letterpaper=true,colorlinks,bookmarks=false]{hyperref}
\usepackage{algorithm} 
\usepackage{chlalgorithmic} 
\usepackage{diagbox}
\usepackage{multirow}

\newcommand{\method}{flex-tuning\xspace}
\newcommand{\Method}{Flex-tuning\xspace}
\newcommand{\methodsnd}{fast flex-tuning\xspace}
\newcommand{\Methodsnd}{Fast flex-tuning\xspace}
\newcommand{\methodtrd}{even faster flex-tuning\xspace}
\newcommand{\Methodtrd}{Even faster flex-tuning\xspace}

\newcommand{\ftall}{\texttt{ft-all}\xspace}       
\newcommand{\ftss}{\texttt{ft-ss}\xspace}
\newcommand{\ftfc}{\texttt{ft-fc}\xspace}
\newcommand{\ftone}{\texttt{ft-fc} (1)\xspace}
\newcommand{\fttwo}{\texttt{ft-fc} (2)\xspace}
\newcommand{\ftflex}{\texttt{flex}\xspace}

\newcommand{\ftflexfast}{\texttt{fast-flex}\xspace}
\newcommand{\ftflexfaster}{\texttt{faster-flex}\xspace}
\newcommand{\cosdist}{$\texttt{cos}_{\mbox{\tiny src}}$}
\newcommand{\cpcdist}{$\texttt{cos}_{\mbox{\tiny pc}}$}
\newcommand{\ourdist}{$\texttt{cos}_{\mbox{\tiny tgt}}$}
\newcommand{\topacc}[1]{\texttt{top-#1}}
\newcommand{\ipc}[1]{\texttt{ipc}}
\newcommand{\comment}[1]{}
\newcommand{\dshift}[2]{\textit{#1}$\rightarrow$ \textit{#2}}
\newcommand{\tinytt}[1]{\texttt{\tiny #1}}
\newcommand{\smalltt}[1]{\texttt{\small #1}}
\newcommand{\idd}{iid}
\newcommand{\D}{\mathcal{D}}

\DeclareMathOperator*{\argmax}{arg\,max}
\newcommand{\IfThenElse}[3]{
  \algorithmicif\ #1\ \algorithmicthen\ #2\ \algorithmicelse\ #3}
\usepackage{makecell}
\definecolor{best}{rgb}{1.0, 0.55, 0.26}
\definecolor{sndbest}{rgb}{0.18, 0.38, 0.66}
\definecolor{links}{rgb}{0.16, 0.36, 0.69}
\definecolor{citations}{rgb}{0.9,0.3,0.2}
\hypersetup{
  colorlinks,
  linkcolor=links,
  pdfborder={0 0 0},
  citecolor=citations
}

\def \thumbheight {10pt}
\def \raiseboxheight{2.5pt}
\newcommand{\thumb}[1]{\hfill\raisebox{-\raiseboxheight}{#1{\thumbheight}}}

\newcommand\copyrighttext{%
\scriptsize Accepted to be Published in: Proceedings of the 2020 IEEE Winter Conference on Applications of Computer Vision, March 2-5, 2020, Snowmass Village, CO, USA
 
2020 IEEE. Personal use of this material is permitted. Permission from IEEE must be obtained for all other uses, in any current or future media, including reprinting/republishing this material for advertising or promotional purposes, creating new collective works, for resale or redistribution to servers or lists, or reuse of any copyrighted component of this work in other works.”}
\newcommand\copyrightnotice{%
\begin{tikzpicture}[remember picture,overlay]
\node[anchor=south west, xshift=45pt, yshift=24pt] at (current page.south west) {\fbox{\parbox{\dimexpr\textwidth-\fboxsep-\fboxrule\relax}{\copyrighttext}}};
\end{tikzpicture}%
}

\ifwacvfinal\fi
\begin{document}

\title{A Flexible  Selection Scheme for Minimum-Effort Transfer Learning}

\author{
  Am\'{e}lie Royer\\
  IST Austria\\
  {\tt\small aroyer@ist.ac.at}
  \and
  Christoph H. Lampert\\
  IST Austria\\
  {\tt\small chl@ist.ac.at}
}

\maketitle
\ifwacvfinal\thispagestyle{empty}\fi

\begin{abstract}
  \textbf{Fine-tuning} is a popular way of exploiting  knowledge contained in a pre-trained
  convolutional network for a new visual recognition task. 
  %
%
  However, the orthogonal setting of transferring knowledge from a pretrained network to a visually different yet semantically close source is rarely considered: This commonly happens with real-life data, which  is not necessarily as clean as the training source (noise, geometric transformations, different modalities, etc.). 

  To tackle such scenarios, we introduce a new, generalized form of fine-tuning, called
  \textbf{flex-tuning}, in which any individual \textbf{unit} (\eg layer) 
  of a network can be tuned, and the most promising one is chosen automatically.
  In order to make the method appealing for practical use, we  propose two
 \textbf{ lightweight and faster selection procedures} that prove to be good approximations  in practice. 
  We study these selection criteria empirically across a variety of domain shifts
  and data scarcity scenarios, and show that fine-tuning individual units, despite its 
  simplicity, yields very good results as an adaptation 
  technique. 
  As it turns out, in contrast to common practice, rather than the last fully-connected unit it is
  best to tune an intermediate or early one in many domain shift scenarios, which is
  accurately detected by flex-tuning.
\end{abstract}
\copyrightnotice

\vspace{-0.32cm}
\section{Introduction}
\label{sec:intro}
Deep  convolutional networks  
have substantially advanced the state of the art in many areas of 
computer vision. 
%
These networks are often interpreted as a 
feature extraction 
stage (typically convolutional layers), followed by a 
 small classifier (fully connected layers),
 and have the ability to learn features from data directly instead
 of having them hard-coded, as was
the case for previous shallow techniques. 
%
However, this comes with a cost, as it 
requires a lot more training data than methods 
relying on fixed ad-hoc feature extraction. 
Consequently, it is not surprising that the first successes of 
deep networks in image classification occurred  as 
large annotated datasets were made available, \eg 
MNIST~\cite{mnist} for digit recognition (60,000 
training samples) or ImageNet~\cite{ILSVRC15} for object 
classification (1.2 million). 

When only little available training data is available, however,
training a deep feature extraction pipeline from scratch
is not possible, as it often 
leads to severe overfitting. 
Instead, two main \emph{transfer learning} strategies have emerged, 
 exploiting the fact that deep convolutional networks 
pre-trained on large datasets are  freely available these
 days~\cite{modelzoo, tensorflowhub, tensornets}:   
\textit{Either}, one isolates and ``freezes'' the feature extraction stage of
the pre-trained model and  then 
uses the available new data to train only the smaller,
 less prone to overfitting, classifier stage, 
\textit{or alternatively}, 
one  fully fine-tunes the model, \ie initializes
the network parameters from the pre-trained network, 
and then trains all layers using the new data, 
typically only for a few steps, to avoid overfitting. 
%
%
%
Choosing the best solution depends not only on the \textit{amount of 
available samples}, though, but also on the \textit{data 
characteristics}.
For example, it has been observed that features learned 
on large and varied natural images datasets, \eg  ImageNet, 
transfer well to related domains such as aerial or even 
biomedical images~\cite{imagenettransfer}. 
However, for domains with very  
different low-level image statistics, \eg  sketches,
fine-tuning all layers is preferable~\cite{phototosketch}. 
Moreover, fine-tuning only a few classification layers is often easier,
hence when both options are viable, one might 
prefer this alternative.

In this work, we argue for a more systematic approach to exploiting 
pre-trained networks, in situations where the new  input domain can \textit{vary 
  greatly in terms of visual appearance}, but its output space shares \textit{similar semantics}
with the one the model was pre-trained on.
%
%
We introduce the idea of \emph{\method{}},
a general-purpose transfer learning scheme that leverages 
the information of an available pre-trained model by fine-tuning 
a \textit{targeted part of the model}, not necessarily the last layer 
or all layers, \textit{but any individual layer or block of consecutive 
layers, selected in a data-dependent way}.
%
In fact, the idea of focusing training resources on specific intermediate layers  
 draws inspiration from an important 
transfer learning paradigm:  
It has been consistently observed across various 
networks and datasets in the literature that early convolutional 
layers capture elementary local properties of images such as edges or local  
textures, while middle layers rather represent configurations 
of several such elements, and the last feature layers extract 
information about high-level concepts, such as object parts and 
their configurations~\cite{Cadena2018DiverseFV, deepdreams, zeilerfergus}.
Thus, in order to adapt, for instance, a network trained on clean natural 
images to work with noisy ones, we hypothesize it is easier to fine-tune 
early layers, while for adapting the same network 
to artistic paintings, focusing on a later layer would be more promising. 

Our contribution is three-fold: \textit{First}, we formally define \emph{\method{}}, which is a strategy for,
given a pre-trained network and a new training dataset, deciding in a data-dependent and automatic way which 
of the available layers to fine-tune, based on a selection criterion on a held-out validation dataset. 
%
\textit{Second}, in order to make \method{} more appealing for practical use,
 we further introduce two variants based on a more efficient selection 
criterion, called \emph{\methodsnd} and \emph{\methodtrd}, that avoid the need to train 
multiple fine-tuned models for the selection process.  
\textit{Finally}, we design an extensive experimental setup that covers varied visual domain shifts,
data scarcity scenarios and architectures. 
We show that \method almost 
always improves classification accuracy over standard fine-tuning techniques,
particularly in settings 
where fine-tuning all layers is prone to overfitting, such as settings with 
small sample size and large networks.
Furthermore, the (even) faster flex-tuning variants are generally on par with \method{} while
providing a much lighter selection procedure.

%
%
%


\section{Related work}
\label{sec:related}
Transferrability of pretrained convolutional networks across 
visual tasks has been often 
observed and extensively studied  in the 
computer vision literature~\cite{DBLP:journals/corr/AzizpourRSMC14, DBLP:conf/eccv/ChuMBHD16, Donahue:2014:DDC:3044805.3044879, Yosinski:2014:TFD:2969033.2969197, zhang2018unreasonable}.
In fact, many state-of-the-art computer vision models are not trained
from random initialization, but rely crucially
on the re-use of weights from networks pre-trained on 
large classification tasks, such as ImageNet~\cite{ILSVRC15}. 
Popular examples include the YOLO object detector~\cite{yolo} 
or  fully-convolutional networks for segmentation~\cite{fcn}. 
%
In the weakly supervised learning literature, pre-trained features
are also used as a compact and semantically 
meaningful image representation, \eg for image retrieval~\cite{Babenko2015AggregatingLD}, 
style transfer~\cite{gatys, johnson}, colorization~\cite{larsson2016learning}, 
or unsupervised part detection~\cite{Simon2015NeuralAC}.
All of these approaches typically aim at transferring knowledge 
between two tasks that have different output structures but similar 
input domain appearances and distributions. 
Closest to our work is \cite{Yosinski:2014:TFD:2969033.2969197}, which studies the outcome of fine-tuning from different levels of a pre-trained network for the standard transfer learning setting. In comparison, we analyze the effect of tuning a single unit of a pre-trained network, in particular for situations where source and target domains are visually dissimilar but semantically close.
%

In fact, our interest lies exactly in these orthogonal scenarios, \ie where one has a similar  
output task, typically multi-class classification, but with 
--potentially significantly -- different \textit{source} and \textit{target}
 input distributions.
This setting resembles, yet differs from, the problem of \emph{domain adaptation}~\cite{saenko2010adapting,gopalan2011domain,Ganin:2016:DTN:2946645.2946704},
where the goal is to construct a classifier for a, usually unlabelled, target task by 
exploiting one or more source tasks. 
In domain adaptation one typically assumes that samples 
from both source and target domain are available, while in the
fine-tuning situation, one only has access to a pre-trained network, 
not the data distribution it was trained on: 
This aspect rules out adversarial training~\cite{discogan, CycleGAN2017}, 
paired samples~\cite{pix2pix}, or more generally, exploiting 
any concrete knowledge from the source distribution to improve 
predictions on the target domain. 

In fact,  with the growth of datasets and necessary compute resources,
the ability to tune networks without access to the original training data is becoming more and more important:
First,  when dealing with very large source datasets,
training jointly on the source and target domains
(as many domain adaptation methods require) is
computationally impractical.
Second, source training data is sometimes non-public,
especially in commercial settings. 
%
Third, specific applications require data privacy, preventing public data release,
for instance for protecting individuals identities in face recognition models. As such, learning under privacy
constraints has become a popular topic in recent years~\cite{private}.

Recent work has also tackled the problem of domain adaptation by transferring from source to 
target directly at the pixel level, either via generative 
models~\cite{45924} or by identifying simpler
causal transformations~\cite{ParKilRojSch18}.
Weight tuning methods are nonetheless simpler to use, as they directly
act on feature representations, rather than learning a transformation
that holds independently of the pre-trained network.

%
%


\section{Flex-Tuning}
\label{sec:proposed}
Our first contribution in this work is to highlight that 
\emph{simple and lightweight, but surprisingly effective, model 
adaptation is possible by fine-tuning the weights of only a 
single unit in a pretrained network, provided that the right unit 
is chosen.}
Which is the right unit depends crucially, and in a 
non-trivial way, on the relation between source and 
target domains as well as on the amount of available 
data. 
We propose to identify the best unit automatically in a 
data-dependent manner using a procedure we call \emph{\method}.

\subsection{Transferring knowledge across domain shifts}
First, we formally introduce the transfer learning  
scenario we are interested in:   
We  are given a pre-trained convolutional network, $N$, 
mapping input space  $X$ to an output space $Y$,
and whose weights were pre-trained on 
a training dataset from a \textit{source domain}, that is however 
not available anymore.
Our goal is to learn a network for a  \textit{target domain},
for which a new, and potentially small, annotated dataset,
$\mathcal{D}\stackrel{\idd}{\sim}\mathcal{P}(X,Y)$, is available. 
In contrast to the standard \textit{transfer learning} application scenario,  
we consider practical settings where the target domain is 
\textit{semantically} close but \textit{visually} different from the source 
domain.
Here, by semantically close, we mean that the output space of the target task is a subset
  of the source task.
Extending the framework to different output structures, \eg from a classification 
 task to a detection task, would be possible  by 
fine-tuning both the unit selected by \method{} and the last fully-connected layer.
In this work, we focus on thoroughly analyzing and characterizing the influence of 
\emph{single units} on transferring knowledge across visually different domains 
and leave the possibility of combining multiple units for future work.

Nonetheless, the setting we consider encompasses a variety of real-world scenarios, 
where the source and target domains do not overlap well.
 For example, we can consider a source network 
 trained on natural images, with the target task of 
classifying monochrome sketches; 
or a source network trained on scenes under  daylight,
 that should also operate at night, \etc.
Here we work with images as inputs, 
and discrete labels as outputs. However, the 
underlying principles apply equally to other input domains and tasks.
%
%


\subsection{\Method}
We consider pre-trained multi-layer convolutional architectures, 
that we decompose into smaller \emph{units}, which 
we denote by $N = N_L \circ \dots\circ N_1$.
In practice, a unit can simply be a single convolutional or fully-connected 
layer, or, for more complex architectures, a block of consecutive 
layers.
Intuitively, we think of units $N_1$ to $N_{L-1}$ as the feature 
extraction part, while the last layer $N_L$ is the performs 
the actual classification, however the method applies to arbitrary decompositions.
Given such a decomposition, the goal of \method is to analyze
 the influence of tuning specific units, 
 not only the last one, for transferring knowledge across domains with different
 visual appearances.
\hyperref[alg:flex]{Algorithm~\ref{alg:flex}} describes the steps of \method 
in pseudo-code: 
For each unit of the network, we construct a fine-tuned network 
$N_{\texttt{ft-$\ell$}}$ by training the network on the available 
target data, allowing only the weights of the $\ell$-th unit to 
change, while keeping all the others frozen. 
We also create a network $N_{\texttt{ft-all}}$, for which 
all layers are fine-tuned. 
We train each network with an early stopping 
criterion, monitoring its performance on the validation set, $\D_{\text{val}}$. 
This prevents overfitting in a way that is data-dependent 
and adaptive to each training setting. In fact, different units 
might have very different numbers of weight parameters, and therefore  
 will often need different numbers of epochs to converge. 
Finally, we choose the best model out of these $L + 1$ networks by comparing their accuracy 
on the validation set and output it as the \emph{flex-tuned} model, 
$N_{\texttt{flex}}$. 

\begin{algorithm}[tb]
\begin{algorithmic}[1]
\INPUT target training and validation sets, $ \D_{\text{train}}$ and $\D_{\text{val}}$
\INPUT pre-trained network with $L$ units, $N=N_L\circ\dots\circ N_1$ 

\smallskip
\FOR{$\ell=1,\dots,L$}
\STATE $N_{\tinytt{ft-$\ell$}} \leftarrow$ fine-tune unit $\ell$ of $N$ on $\D_{\text{train}}$ until

                      \quad\ \ \,\qquad accuracy on $\D_{\text{val}}$ stops improving

\STATE $a_{\tinytt{ft-$\ell$}} \leftarrow$ accuracy  of $N_{\tinytt{ft-$\ell$}}$ on $\D_{\text{val}}$
\ENDFOR
\STATE $N_{\tinytt{ft-all}} \leftarrow$ fine-tune all units of $N$ on $\D_{\text{train}}$ until

                      \qquad\ \qquad accuracy on $\D_{\text{val}}$ stops improving
                      
\STATE $a_{\tinytt{ft-all}} \leftarrow$ accuracy  of $N_{\tinytt{ft-all}}$ on $\D_{\text{val}}$

\smallskip 
\STATE $N_{\tinytt{flex}} \leftarrow N_{\tinytt{best}}$\quad for\quad  $\smalltt{best} \leftarrow \!\!\!\!\argmax\limits_{X\in\{\texttt{1,\dots,L,all}\}}\!\!\!\! a_{\tinytt{ft-X}}$

\smallskip 
\OUTPUT $N_{\ftflex{}}$
\end{algorithmic}
\caption{Flex-Tuning (\texttt{flex})}\label{alg:flex}
\end{algorithm}

\begin{figure}[t]
  \includegraphics[width=\linewidth]{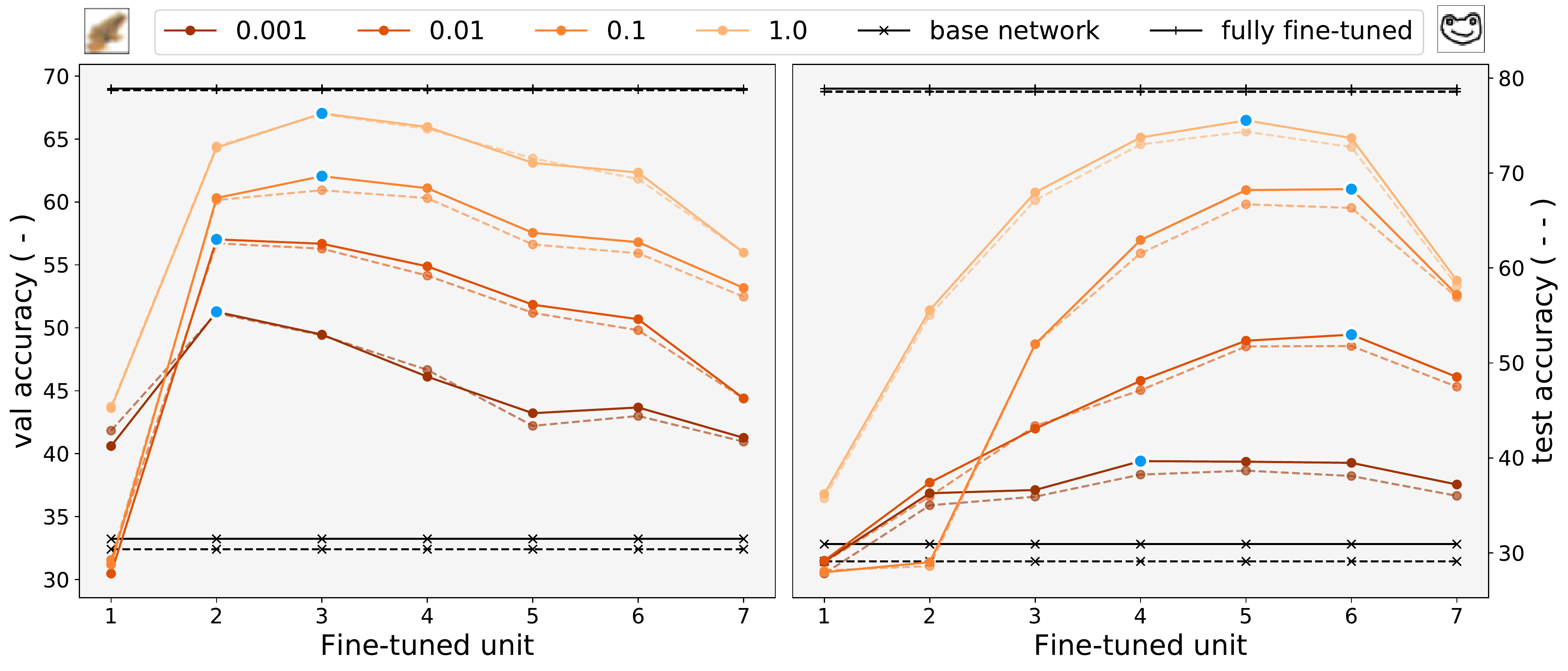}
  \vspace{-0.4cm}
  \caption{Validation (-) and test (- -) accuracies for  fine-tuning a single unit
    of a pretrained CIFAR network to the Blurry CIFAR (left) and Quick, Draw! (right) datasets. Each line color
    represents a different subsampling ratio of the target training dataset, while blue markers indicate the unit
    picked based on validation accuracy.}\label{fig:teaser}
\end{figure}

\subsection{Practicality of the method}
Technically, \hyperref[alg:flex]{Algorithm~\ref{alg:flex}} performs 
an exhaustive search over the potential fine-tuned models.
Therefore, the existing theoretical results for 
model selection~\cite{shalev2014understanding} apply, 
and we obtain that, in the limit\comment{of infinite data}, \method 
will indeed choose the best of the possible models. Moreover, the 
difference between \method's accuracy estimated from the 
validation data and the expected accuracy on future data 
decreases with a rate of $O\big(\sqrt{\frac{\log L}{|\D_{\text{val}}|}}\big)$.
%
%
%
In  \hyperref[fig:teaser]{Figure \ref{fig:teaser}}, we illustrate \method's practical use: 
We apply the proposed method on a small network (5 convolutional and 2 fully connected layers) 
pre-trained on CIFAR~\cite{cifar} and to be adapted to a subset of the ``Quick, Draw!'' 
dataset~\cite{quickdraw} and a blurred variant of CIFAR, for different sizes 
of the target training dataset.
These preliminary results  show that (i) it is often beneficial to fine-tune 
an intermediate layer rather than the last one and that 
(ii) well-performing units strongly depend on the dataset and in a non-trivial way,
 but can be efficiently pin-pointed with a simple
selection criterion such as \method.

For deep networks however, \method{} can be 
computationally costly: It requires 
training as many networks as there are units, plus 
another one in which all units are fine-tuned. 
Let us denote the average number of training epochs 
by $E_{\text{one}}$ when fine-tuning a single unit, 
and by $E_{\text{all}}$ when fine-tuning all. 
Also, let us denote the corresponding average computational cost of 
one such epoch as $c_{\text{one}}$ and $c_{\text{all}}$, 
respectively.
 Then the total runtime complexity of \method{} 
is $O(LE_{\text{one}}c_{\text{one}} + E_{\text{all}}c_{\text{all}})$.
Even when taking into account that typically $E_{\text{all}}>E_{\text{one}}$ 
and $c_{\text{all}}> c_{\text{one}}$, for reasonably 
large networks the complexity is often dominated by
the computational cost of fine-tuning the network
 once for each unit. 
Since ultimately only one of the models 
is chosen, these computations end up wasted.
To address this issue, we introduce two improved selection criteria in the
following section to efficiently approximate \method.

\section{Efficient Selection Criteria}
\label{sec:fasterproposed}

\subsection{\Methodsnd} 
To overcome the aforementioned computational inefficiency of \method,
we propose a different criterion, \emph{\methodsnd{}}, for 
selecting the unit to be fine-tuned.
It relies on the idea that a given unit's influence can be approximated by \textit{a few
  feed-forward passes} rather than a full training process.
While it does not come with formal guarantees, we found it to work 
nearly as well as the exhaustive search in practice, while at the
same time requiring only $2$ networks to be trained instead of $L+1$. 
\hyperref[alg:fastflex]{Algorithm~\ref{alg:fastflex}} describes \methodsnd{} in pseudo-code: 
 The method starts by training one new model, $N_{\tinytt{ft-all}}$,
by fine-tuning all units of the pre-trained network on the training
data available for the target domain. 
From this, we construct $L$ new networks by \emph{network surgery}. 
For any $\ell=1,\dots,L$, we create a proxy network, $N_{\tinytt{prox-}\text{\tiny $\ell$}}$, 
by copying all units from $N$, except the $\ell$-th one, 
which is copied from the fine-tuned network, $N_{\tinytt{ft-all}}$.
Clearly, the resulting hybrid networks are not functional models, as 
the $l$-th unit and the other units were not trained together.
Nevertheless, the construction allows us to derive a measure 
which of the network units is the most promising candidate 
for fine-tuning, namely the one that leads to the\textit{ biggest 
improvement in accuracy (if any) when applied to the target domain}.
Numerically, we  compute the accuracy of each model $N_{\tinytt{prox-}\text{\tiny $\ell$}}$
on the validation dataset and identify the value for $\ell$ with highest 
accuracy. 
We then create a viable model by fine-tuning the selected unit on the target dataset $\D$.
Finally, we output either this model, or the one in which all layers
were fine-tuned (which is  available as we created it at the
beginning of the procedure), depending on which achieved the higher
validation accuracy. 
%
We report the validation accuracies of the $N_{\tinytt{prox-}\text{\tiny $\ell$}}$ models for our different experimental settings in the supplemental material.

In comparison to \method, \methodsnd only has to fine-tune two networks instead of $L+1$. 
Its runtime complexity is hence $O(E_{\text{one}}c_{\text{one}} + E_{\text{all}}c_{\text{all}})$,
thereby providing substantial computational savings for large
networks.

\begin{algorithm}[tb]
\begin{algorithmic}[1]
\INPUT target training and validation sets, $ \D_{\text{train}}$ and $\D_{\text{val}}$
\INPUT pre-trained network with $L$ units, $N=N_L\circ\dots\circ N_1$ 

\smallskip
\STATE $N_{\tinytt{ft-all}} \leftarrow$ fine-tune all units of $N$ on $\D_{\text{train}}$ until

                      \qquad\,\qquad accuracy on $\D_{\text{val}}$ stops improving
\STATE $a_{\tinytt{ft-all}} \leftarrow$ accuracy  of $N_{\tinytt{ft-all}}$ on $\D_{\text{val}}$

\smallskip
\FOR{$\ell=1,\dots,L$}

\STATE $N_{\tinytt{prox-}\text{\tiny $\ell$}} \leftarrow N_{L}\circ\dots\circ N_{l+1}\circ [N_{\tinytt{ft-all}}]_{\ell}\circ N_{\ell-1}\circ\dots\circ N_{1}$

\STATE $a_{\ell} \leftarrow$ accuracy  of $N_{\tinytt{prox-}\text{\tiny $\ell$}}$ on $\D_{\text{val}}$
\ENDFOR
\STATE $\smalltt{best} \leftarrow \ \argmax_{\ell}\ \ a_{\ell},\ \ell\in\{\texttt{1,\dots,L}\}$

\STATE $N_{\tinytt{best}} \leftarrow$ fine-tune unit \ \smalltt{best} of $N$ on $\D_{\text{train}}$ until

                      \quad\ \ \qquad accuracy on $\D_{\text{val}}$ stops improving

\STATE $a_{\tinytt{ft-best}}  \leftarrow$ accuracy  of $N_{\tinytt{best}} $ on $\D_{\text{val}}$
\STATE $N_{\tinytt{flex}} \leftarrow$ \IfThenElse{$a_{\tinytt{ft-best}} \geq a_{\tinytt{ft-all}}$}{$N_{\tinytt{best}}$}{$N_{\tinytt{ft-all}}$}
\OUTPUT $N_{\tinytt{flex}}$
\end{algorithmic}
\caption{Fast Flex-Tuning (\texttt{fast-flex})}\label{alg:fastflex}
\end{algorithm}

\begin{algorithm}[tb]
\begin{algorithmic}[1]
\INPUT target training and validation sets, $ \D_{\text{train}}$ and $\D_{\text{val}}$
\INPUT pre-trained network with $L$ units, $N=N_L\circ\dots\circ N_1$  

\smallskip
\STATE $N_{\tinytt{ft-all}} \leftarrow$ fine-tune all units of $N$ on $\D_{\text{train}}$ 

                      \qquad\ \qquad for a single epoch

\FOR{$\ell=1,\dots,L$}

\STATE $N_{\tinytt{prox-}\text{\tiny $\ell$}} \leftarrow N_{L}\circ\dots\circ N_{l+1}\circ [N_{\tinytt{ft-all}}]_{\ell}\circ N_{\ell-1}\circ\dots\circ N_{1}$

\STATE $a_{\ell} \leftarrow$ accuracy  of $N_{\tinytt{prox-}\text{\tiny $\ell$}}$ on $\D_{\text{val}}$
\ENDFOR
\STATE $\smalltt{best} \leftarrow \ \argmax_{\ell}\ \ a_{\ell},\ \ell\in\{\texttt{1,\dots,L}\}$

\STATE $N_{\tinytt{flex}} \leftarrow$ fine-tune unit \ \smalltt{best} of $N$ on $\D_{\text{train}}$ until

                      \quad\ \ \qquad accuracy on $\D_{\text{val}}$ stops improving

\OUTPUT $N_{\tinytt{flex}}$
\end{algorithmic}
\caption{Even Faster Flex-Tuning (\texttt{faster-flex})}\label{alg:fasterflex}
\end{algorithm}


\subsection{\Methodtrd}  

In some situations, training 
from scratch or  fine-tuning the complete network is simply 
computationally too costly: Neither \method nor \methodsnd are applicable, 
as both require  training a network by fine-tuning 
all units as the first step of their selection process. 
To overcome this, we propose an even faster variant, as described
in \hyperref[alg:fasterflex]{Algorithm~\ref{alg:fasterflex}}.

\emph{\Methodtrd} resembles \methodsnd in that it 
selects a unit to be fine-tuned based on the accuracies 
of different proxy models that are obtained by network 
surgery, each time preserving $L-1$ units from the pre-trained source network
 and replacing the remaining one with its fine-tuned counterpart.
The difference lies in that the fine-tuned units 
are obtained from a network in which all 
units have been fine-tuned for just a \emph{single epoch}. 
This results in a total computational runtime of 
$O(E_{\text{one}}c_{\text{one}} + c_{\text{all}})$.
We consider this close to optimal for an adaptive
technique, as at least the cost $E_{\text{one}}c_{\text{one}}$ 
clearly cannot be avoided, if the goal is to produce a 
network in which at least one unit has been fine-tuned. 
The drawback of the acceleration is that the \emph{\methodtrd} 
algorithm does not have access to a reliable estimate of what 
performance a network with all units fully fine-tuned would have achieved.
This is however not relevant here as, by assumption, 
the computational budget does not suffice for training 
such a model anyway.

In summary, \methodtrd  is  a generalization of fine-tuning the last
  unit of the network, as is often done in practice, but instead the
most promising unit  is  chosen by  a brief  selection process. 
\hyperref[tab:runtime]{Table~\ref{tab:runtime}} summarizes the runtime complexity
of all proposed models, as well as the two  main 
 baselines we use in our experiments: 
\texttt{ft-fc}, which fine-tunes always the last unit 
(\ie the fully-connected layer(s)), and \texttt{ft-all}, 
which fine-tunes always all layers.

\begin{table}[t]
  \centering
\begin{tabular}{|r|c|} 
\hline
 method & computational cost  
\\\hline
\texttt{flex} &  $LE_{\text{one}}c_{\text{one}} + E_{\text{all}}c_{\text{all}}$     
\\
\texttt{fast-flex} &  $E_{\text{one}}c_{\text{one}}\  +\ E_{\text{all}}c_{\text{all}}$  
\\
\texttt{faster-flex} & $E_{\text{one}}c_{\text{one}} + c_{\text{all}}$
\\\hline
\texttt{ft-fc} & $E_{\text{one}}c_{\text{one}}$
\\
\texttt{ft-all} & $E_{\text{all}}c_{\text{all}}$
\\\hline
\end{tabular}
  \vspace{0.1cm}
\caption{Runtime complexities.
$L$ is the number of units in the network, $E_{\text{one}}$ 
and $c_{\text{one}}$ are the average number of epochs 
until early stopping for fine-tuning one unit, and the 
estimated cost of one such epoch. 
$E_{\text{all}}$ and $c_{\text{all}}$ are the analogous
quantities when fine-tuning all network units. In general, $E_{\text{all}}>E_{\text{one}}$ 
and $c_{\text{all}}> c_{\text{one}}$. }\label{tab:runtime} 
\end{table}
 	


\section{Experiments}
\label{sec:experiments}
In this section, we introduce our experimental setup, covering a large number of domain shifts and data scarcity scenarios. 
We then describe  fine-tuning baselines commonly used in the literature, and compare them  to the proposed methods, \method{} ({\small \ftflex{}}),  \methodsnd ({\small \ftflexfast{}}) and \methodtrd{} ({\small \ftflexfaster{}}).

\subsection{Experimental set-up}
\label{sec:datasets}
We build several domain shift scenarios, 
ranging from simple parametric transformations to severe visual appearance shifts.
In order to explore the impact of data scarcity, we additionally consider several subsampled versions of each target dataset, ranging from a few images per class to hundreds of them.
The different settings are thus mainly characterized by: (i) the depth of the base source network, (ii) the size of the target dataset we tune on, and (iii) the type of input domain shift: simple parametric transformations, \eg   manipulating color channels, complex (non-trivially invertible) parametric transformations, and general free-range transformations.
We summarize our setup in \hyperref[tab:datasets]{Table \ref{tab:datasets}}.

\medskip\noindent\textbf{Medium-sized experiments.}
We first consider a small 4 layers network (which we decompose in 4 one-layer units: 2 convolutional layers followed by 2 fully-connected ones) pretrained on a subset of MNIST training images.
We use the remaining samples (except 5000 of them that we keep for validation) to build synthetic domain shifts such as \emph{affine transformations} (randomized or fixed for all images), or \emph{random occlusions}. 
Second, we build a 7 layers network (7 one-layer units: 5 convolutional and 2 fully connected ones) that we pre-train on half of the CIFAR training set~\cite{cifar}.
As target domains, we consider several synthetic transformations of the remaining samples, as well as a subset of the QuickDraw dataset~\cite{quickdraw}: We restrict ourselves to the object classes they have in common, \ie all CIFAR classes except for ``deer''.
We also consider the converse setting, \ie pre-training on QuickDraw and using as target domains CIFAR and synthetically generated blurry and noisy QuickDraw samples.
Since both aforementioned architectures have two  fully connected layers, we consider two baselines, \ftone{} and \fttwo{}, corresponding to fine-tuning only the last, or the last two fully-connected layers respectively.

\medskip\noindent\textbf{Large-scale setting.}
Finally, we consider two large-scale settings using the Inception2 architecture~\cite{inceptionv2, tensornets, inceptionv1}.
We decompose the model so as to not separate layers belonging to the same Inception module, which results in 13 units, the last one being the  single fully-connected classification layer of the architecture.
We first experiment on synthetic transformations of natural images. For this setting, we use a network pretrained on ILSVRC2012-train. We then split ILSVRC2012-val in three parts. 25k images are used to create target datasets, 5k are kept for validation and the remaining 20k are used for testing.
Second, we consider the more challenging setting of stylistic transformations using the PACS dataset~\cite{pacs}, initially introduced for the task of domain generalization: 
We use art paintings, cartoons and sketches, as target domains, which we further split into train/val/test sets.
In this setting, the target task is a subset of the source ILSVRC classification task (ignoring the ``person'' class in PACS as it does not have an equivalent). 
%
%

%
\def \basedomainwidth {0.13\linewidth}
\def \targetdomainwidth {0.13\linewidth}

\def \mnisttrim {0.95cm 0.70cm 24cm 0.85cm}
\newcommand{\mnistimg}[1]{\includegraphics[trim={0.95cm 0.70cm 24cm 0.85cm}, clip, width=#1]{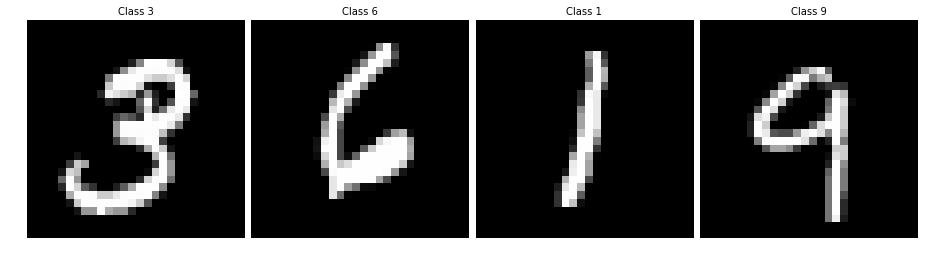}}
\newcommand{\blurrymnistimg}[1]{\includegraphics[trim={16.85cm 0.70cm 8.1cm 0.85cm}, clip, width=#1]{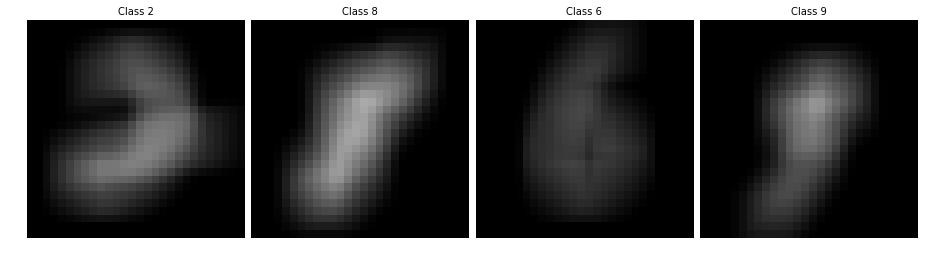}}
\newcommand{\occludedmnistimg}[1]{\includegraphics[trim={16.85cm 0.70cm 8.1cm 0.85cm}, clip, width=#1]{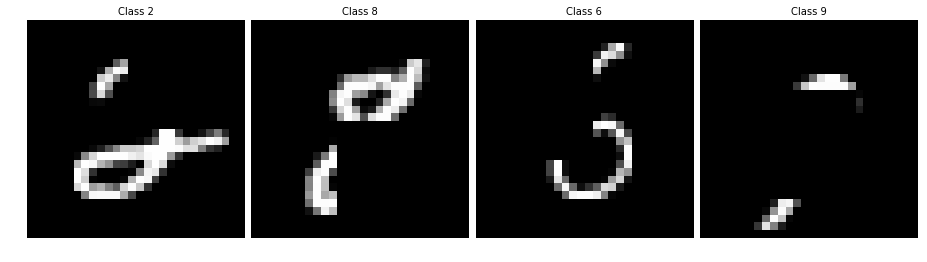}}
\newcommand{\mnistmimg}[1]{\includegraphics[trim={16.85cm 0.70cm 8.1cm 0.85cm}, clip, width=#1]{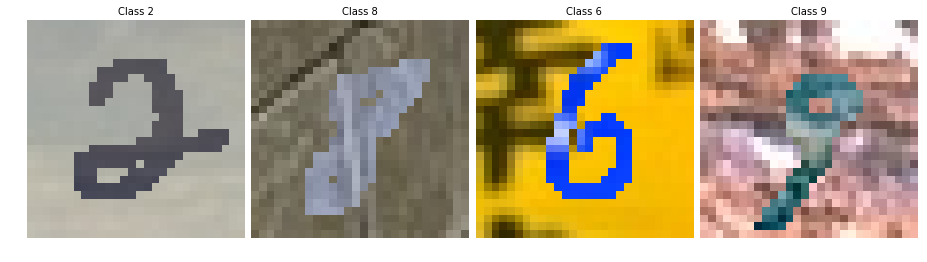}}
\newcommand{\rndtransformmnistimg}[1]{\includegraphics[trim={16.85cm 0.70cm 8.1cm 0.85cm}, clip, width=#1]{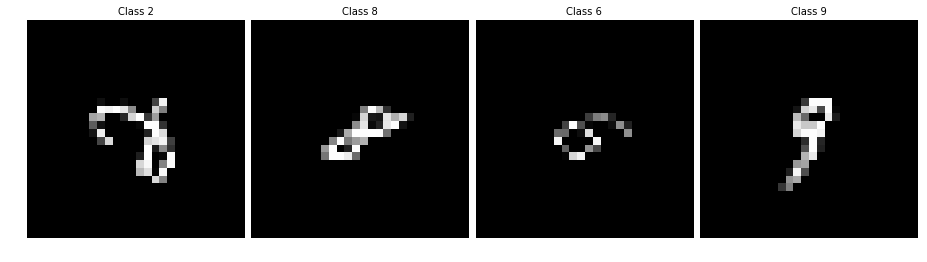}}
\newcommand{\fixtransformmnistimg}[1]{\includegraphics[trim={16.85cm 0.70cm 8.1cm 0.85cm}, clip, width=#1]{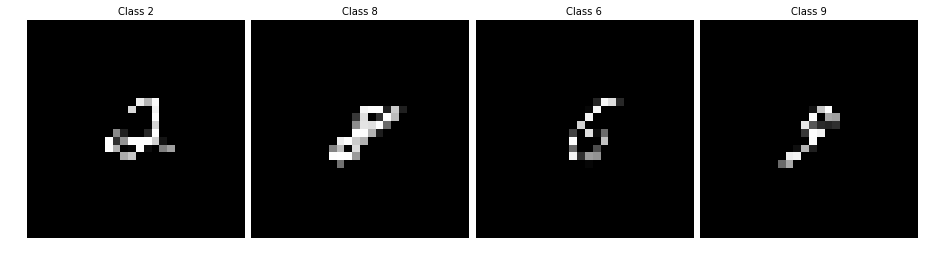}}
\newcommand{\svhnimg}[1]{\includegraphics[trim={8.9cm 0.70cm 16.3cm 0.85cm}, clip, width=#1]{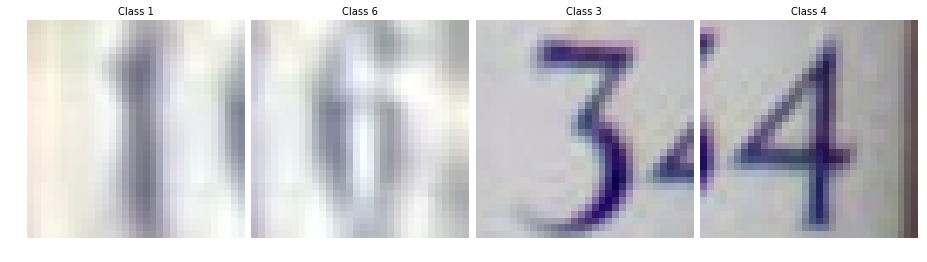}}
\newcommand{\cifarsrcimg}[1]{\includegraphics[trim={16.85cm 0.70cm 8.1cm 0.85cm}, clip, width=#1]{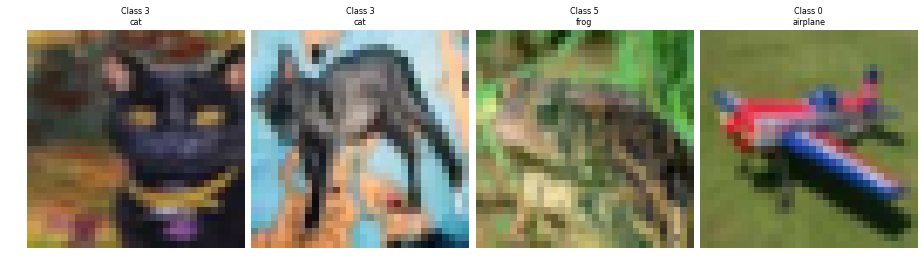}}
\newcommand{\blurrycifarimg}[1]{\includegraphics[trim={24.7cm 0.70cm 0.42cm 0.85cm}, clip, width=#1]{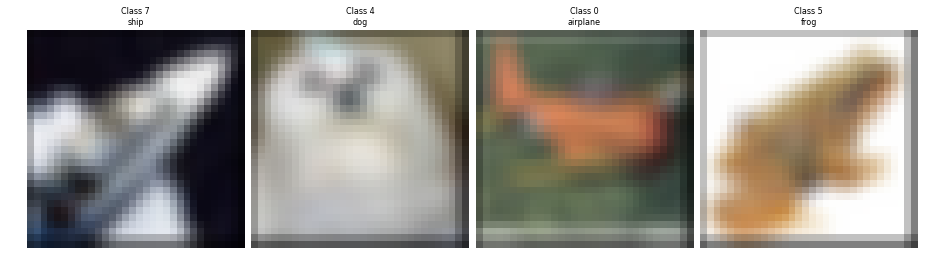}}
\newcommand{\noisycifarimg}[1]{\includegraphics[trim={24.7cm 0.70cm 0.42cm 0.85cm}, clip, width=#1]{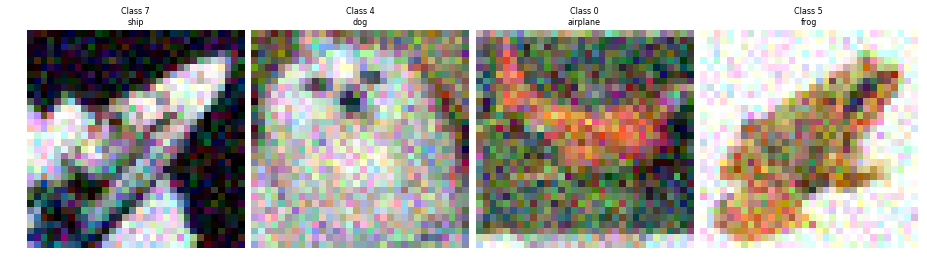}}
\newcommand{\quickdrawimg}[1]{\includegraphics[trim={8.9cm 0.70cm 16.3cm 0.85cm}, clip, width=#1]{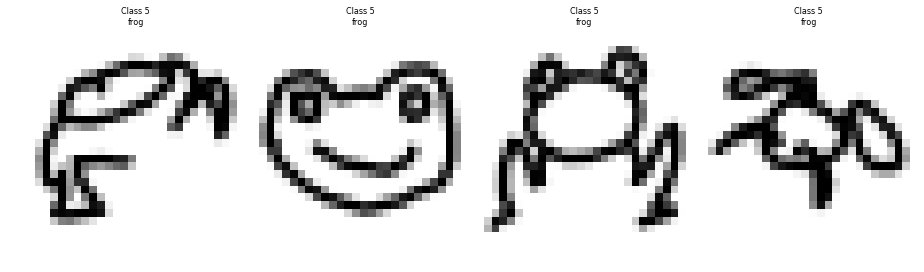}}
\newcommand{\quickdrawsrcimg}[1]{\includegraphics[trim={24.7cm 0.70cm 0.42cm 0.85cm}, clip, width=#1]{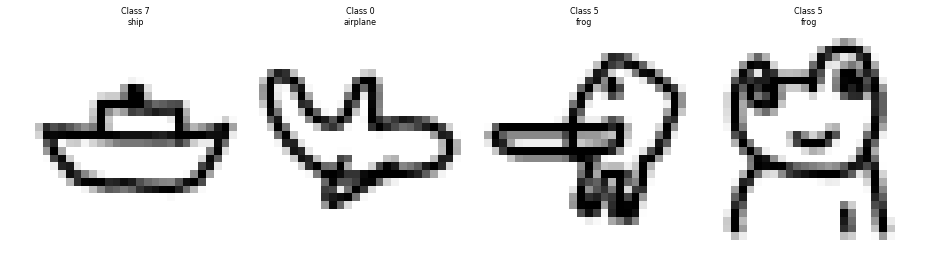}}
\newcommand{\blurryquickdrawimg}[1]{\includegraphics[trim={8.9cm 0.70cm 16.3cm 0.85cm}, clip, width=#1]{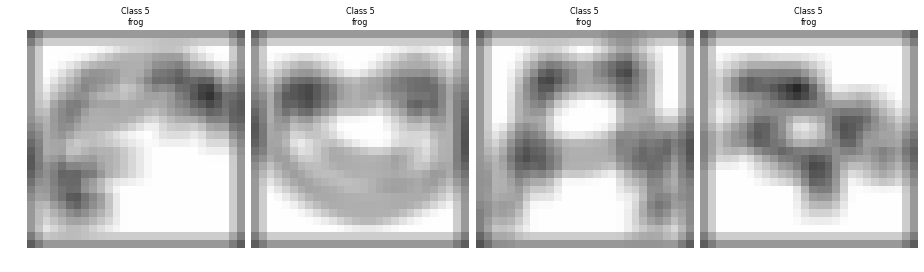}}
\newcommand{\noisyquickdrawimg}[1]{\includegraphics[trim={8.9cm 0.70cm 16.3cm 0.85cm}, clip, width=#1]{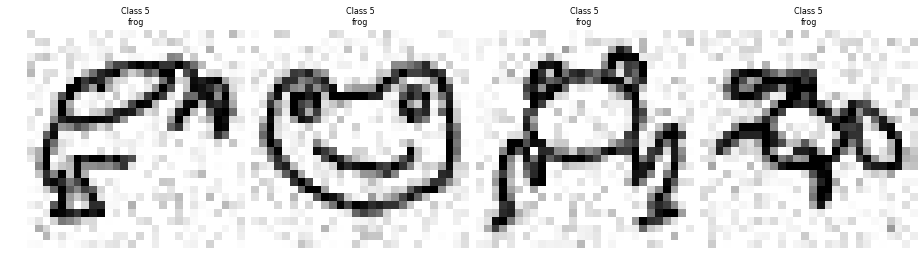}}
\newcommand{\cifarimg}[1]{\includegraphics[trim={24.7cm 0.70cm 0.42cm 0.85cm}, clip, width=#1]{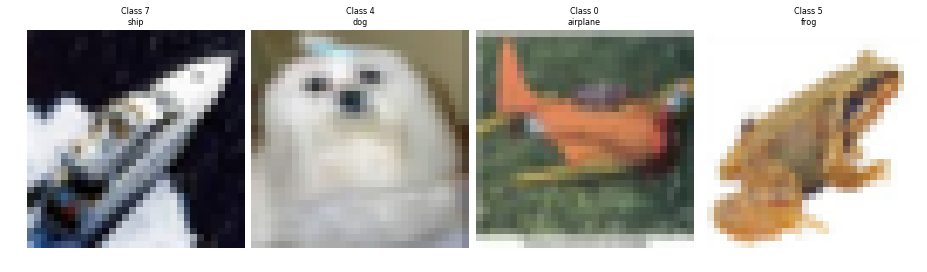}}
\newcommand{\ilsvrcimg}[1]{\includegraphics[width=#1]{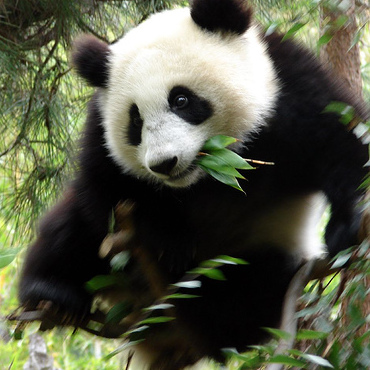}}
\newcommand{\yuvimg}[1]{\includegraphics[trim={8.9cm 0.70cm 16.3cm 0.85cm}, clip, width=#1]{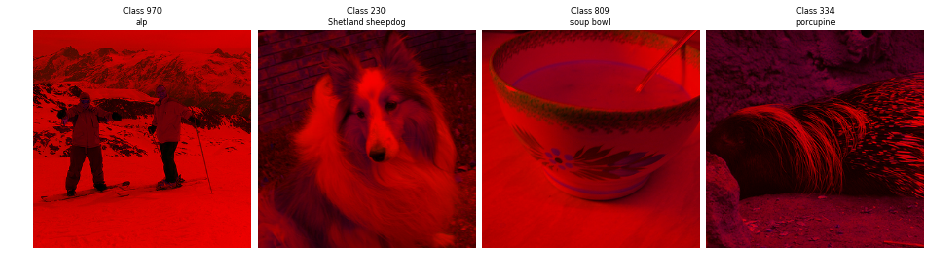}}
\newcommand{\hsvimg}[1]{\includegraphics[trim={8.9cm 0.70cm 16.3cm 0.85cm}, clip, width=#1]{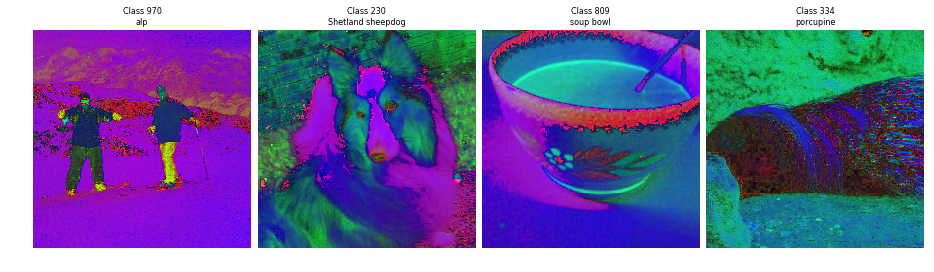}}
\newcommand{\fixedkaleidoimg}[1]{\includegraphics[trim={8.9cm 0.70cm 16.3cm 0.85cm}, clip, width=#1]{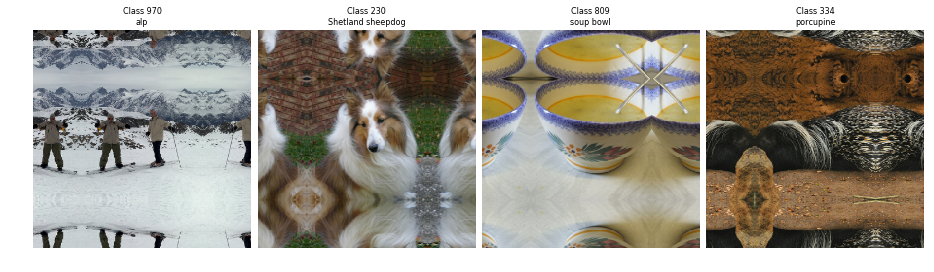}}
\newcommand{\fixedscaleimg}[1]{\includegraphics[trim={8.9cm 0.70cm 16.3cm 0.85cm}, clip, width=#1]{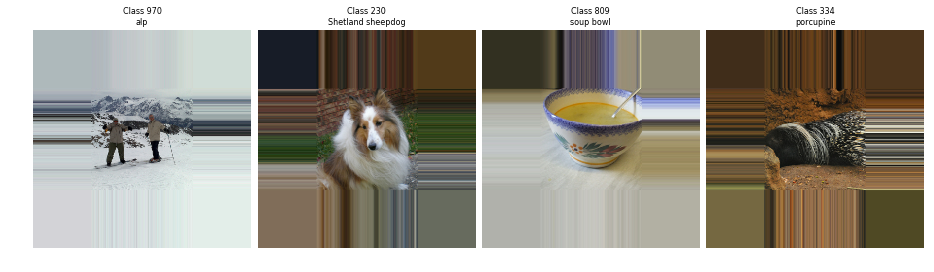}}
\newcommand{\fixedrotationimg}[1]{\includegraphics[trim={8.9cm 0.70cm 16.3cm 0.85cm}, clip, width=#1]{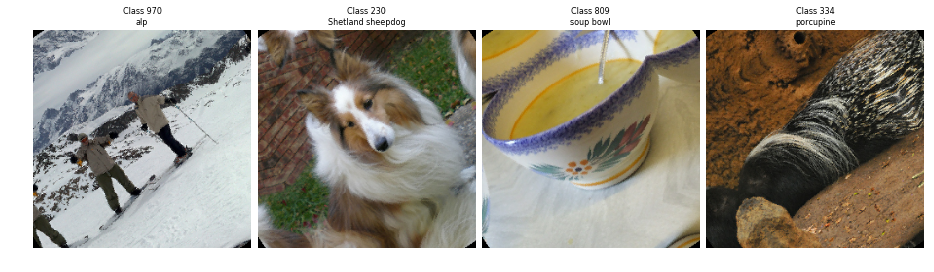}}
\newcommand{\photoimg}[1]{\includegraphics[trim={0.95cm 0.70cm 24cm 0.85cm}, clip, width=#1]{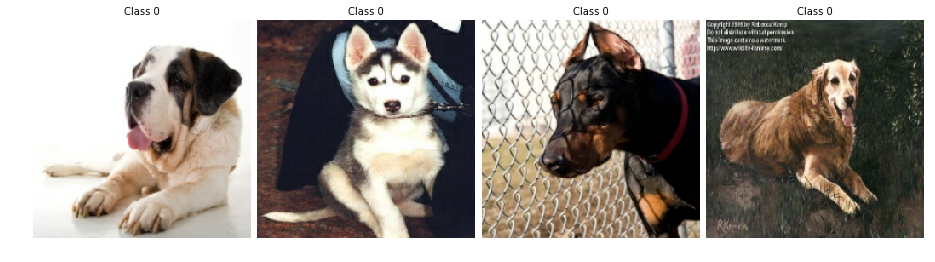}}
\newcommand{\artimg}[1]{\includegraphics[trim={16.85cm 0.70cm 8.1cm 0.85cm}, clip, width=#1]{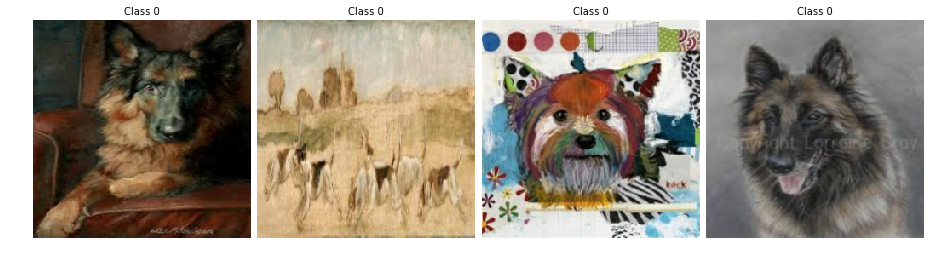}}
\newcommand{\sketchimg}[1]{\includegraphics[trim={8.9cm 0.70cm 16.3cm 0.85cm}, clip, width=#1]{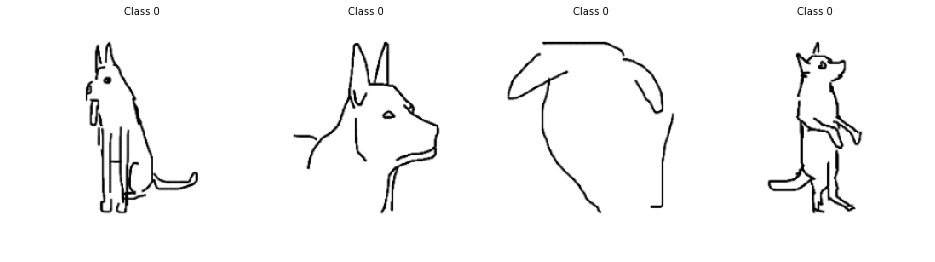}}
\newcommand{\cartoonimg}[1]{\includegraphics[trim={8.9cm 0.70cm 16.3cm 0.85cm}, clip, width=#1]{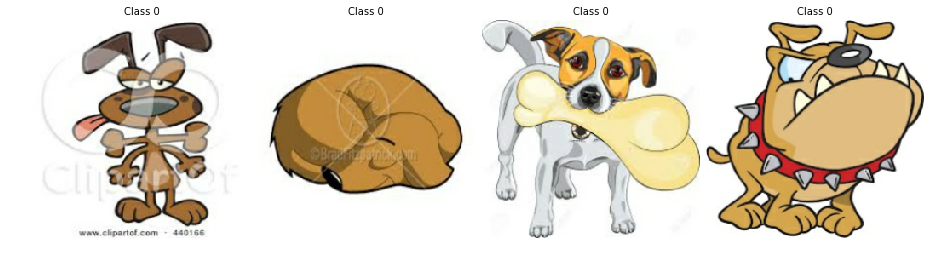}}

\begin{table}[!tb]  
  \renewcommand\cellset{\renewcommand\arraystretch{0.805}%
\setlength\extrarowheight{0pt}}
  \footnotesize
  \centering  
  \resizebox{\linewidth}{!}{%
  \begin{tabular}{|c|c|}
    \hline
    \normalsize Source & \normalsize Target domains \\
    \hline
    \makecell{
    \mnistimg{\basedomainwidth}\\
    \textbf{MNIST {\tiny (subset)}}~\cite{mnist}\\
    25k images\\
    10 classes\\
    4-layers \\
    \topacc{1}:  0.989 }
           &
             \makecell{
             \makecell[t]{
             \blurrymnistimg{\targetdomainwidth} \\
    \textbf{Blurry} \\
    \topacc{1}: 0.748  \\
    }~
    \makecell[t]{
    \occludedmnistimg{\targetdomainwidth}\\
    \textbf{Occluded } \\
    \topacc{1}: 0.581 \\
    }~
    \makecell[t]{
    \mnistmimg{\targetdomainwidth}\\
    \textbf{MNIST-M}~\cite{Ganin:2015:UDA:3045118.3045244} \\
    \topacc{1}: 0.439 \\
    }
    \\
    %
    \makecell[t]{
    \rndtransformmnistimg{\targetdomainwidth} \\
    \textbf{Transform} \\
    \textbf{(random)}\\
    \topacc{1}: 0.322 \\
    }~
    \makecell[t]{
    \svhnimg{\targetdomainwidth} \\
    \\
    \textbf{SVHN}~\cite{Netzer2011ReadingDI} \\
    \topacc{1}: 0.211 \\
    }~
    \makecell[t]{
    \fixtransformmnistimg{\targetdomainwidth} \\
    \textbf{Transform }\\
    \textbf{(fixed)}\\
    \topacc{1}: 0.160 \\
    }
    \\
    \\
    \textbf{ratios} $\sim$ 3, 30, 300 and 3k images per class}
    \\ 
    \hline
    \makecell[t]{
    \cifarsrcimg{\basedomainwidth}\\
    \textbf{CIFAR {\tiny (subset)}}~\cite{cifar}\\
    18k images\\
    9 classes\\
    7-layers\\
    \topacc{1}:  0.738}
           &
             \makecell[t]{
             \makecell[t]{
             \noisycifarimg{\targetdomainwidth} \\
    \textbf{Noisy } \\
    \topacc{1}: 0.540 \\
    }~~
    \makecell[t]{
    \blurrycifarimg{\targetdomainwidth} \\
    \textbf{Blurry}  \\
    \topacc{1}: 0.324 \\
    }~~
    \makecell[t]{
    \quickdrawimg{\targetdomainwidth} \\
    \textbf{QuickDraw}~\cite{quickdraw} \\
    \topacc{1}: 0.291 \\
    }
    \\
    \\
    \\
    \textbf{ratios} $\sim$ 2, 20, 200 and 2k images per class}
    \\
    \hline
    \multirow{ 2}{*}{\makecell{
    \ilsvrcimg{\basedomainwidth}\\
    \textbf{ILSVRC}~\cite{ILSVRC15} \\
    \textbf{('12 train split)}\\
    1M images\\
    1000 classes\\
    Inception2\\
    \topacc{5}:  0.918}}
           &
             \makecell{
             \makecell[t]{
             \yuvimg{\targetdomainwidth} \\
    \textbf{YUV } \\
    \topacc{5}: 0.841 \\
    }
    \makecell[t]{
    \fixedrotationimg{\targetdomainwidth} \\
    \textbf{Fixed rotation } \\
    \topacc{5}: 0.743 \\
    }
    \\
    \makecell[t]{
    \fixedkaleidoimg{\targetdomainwidth} \\
    \textbf{Fixed scaling} \\
    {\scriptsize \textbf{(symmetric pad)}} \\
    \topacc{5}: 0.519 \\
    }~
    %
    \makecell[t]{
    \fixedscaleimg{\targetdomainwidth} \\
    \textbf{Fixed scaling}\\
    {\scriptsize  \textbf{(stretch pad)}} \\
    \topacc{5}: 0.440 \\
    }~
    \makecell[t]{
    \hsvimg{\targetdomainwidth} \\
    \\
    \textbf{HSV } \\
    \topacc{5}: 0.384 \\
    }    
    \\
    \\
    \textbf{ratios} $\sim$ 2, 12 and 25 images per class
    }
    \\
    \cline{2-2}
           &
             \makecell[t]{
             \makecell[t]{
             \artimg{\targetdomainwidth}\\
    \textbf{Art}\\
    \topacc{1}: 0.532  \\
    }~~
    \makecell[t]{
    \cartoonimg{\targetdomainwidth}\\
    \textbf{Cartoon}\\
    \topacc{1}: 0.346 \\
    }~~
    \makecell[t]{
    \sketchimg{\targetdomainwidth}\\
    \textbf{Sketch} \\
    \topacc{1}: 0.142 \\
    }
    \\
    \\
    \textbf{ratios} $\sim$ 2, 20 and 200 images per class}
    \\
    \hline
  \end{tabular}
  }
  \caption{\label{tab:datasets} Source domains and architectures (\emph{left}) we consider, with the corresponding target domains (\emph{right}) and the training dataset \textbf{subsampling ratios} we consider, as the average number of images per class:  the last entry corresponds to the full dataset size.
    %
  }
\end{table}

\medskip\noindent\textbf{Baselines.}
\label{sec:baselines}
We first consider the two most common transfer learning schemes as baselines. Starting with a network initialized with the same weights and architectures as the source pre-trained network, $N$: (i) \ftall{} consists in fine-tuning \textit{all } layers $N_1, \dots N_L$ on the training set $\D$ from the target domain, and (ii) \ftfc{}, which corresponds to fine-tuning only the last \textit{fully-connected }units of the network, while keeping earlier units frozen.
%
 We also consider using scaling and shifting operations as in \cite{scaling} and  refer to this baseline as \ftss{}: It consists in fine-tuning the last classification layer as well as lightweight kernel-scaling and bias-shifting parameters at every layer. Thus \ftss{} acts on all levels of the architecture, but requires few additional learning parameters, hoping to prevent overfitting problems.
%
%
%
%
%
%

\medskip\noindent\textbf{Training.}
%
%
We  measure  performance as \topacc{1} classification accuracy, and \topacc{5} for ILSVRC-based domains.
We use the same hyperparameters as were used during training of the base source network. As is common, for finetuning, we use a lower base learning rate: $10^{-3}$ for the small convolutional networks, and  $10^{-4}$ for the Inception2 networks.
We train all models using the Adam~\cite{adam} optimizer.
As mentioned previously, we also employ an early stopping criterion based on  validation  accuracy, regularly computed during training (every 5-10 epochs).
This also dampens the negative effect of overfitting in scenarios that are overly prone to it (\eg \ftall{} with small sample size and a large network).
Finally, in the very scarce data setting  ($\sim$ 1 image per class) we report metrics averaged over 20 runs, to avoid a potential bias towards the sampled training images.

\subsection{Main results}
\label{sec:results}
\begin{table*}[tb]
  \def \firstcolstep {0.25\tabcolsep}
  \renewcommand{\arraystretch}{1.2}

  %
  \begin{minipage}[c]{0.5\textwidth}
  \resizebox{\linewidth}{!}{%
    \begin{tabular}{|l@{\hspace{ \firstcolstep}}c||c|c|c||c|c|c|}
      \hline
       \multicolumn{2}{|c||}{\multirow{2}{*}{\small \textbf{ILSVRC \thumb{\ilsvrcimg}}}} & \multicolumn{3}{c||}{\small \ftflex{}} & \multicolumn{3}{c|}{\small \texttt{ft-}} \\
      \cline{3-8}
                                                                                               & &  \footnotesize \texttt{flex} & \footnotesize \texttt{fast} & \footnotesize \texttt{faster} & \footnotesize \texttt{fc} & \footnotesize \texttt{ss} & \footnotesize \texttt{all} \\
      \hline
      \hline
      \multicolumn{7}{|l|}{\textbf{ratio:} 2 images per class}  \\
      \hline
      \small Art   & (0.53) \thumb{\artimg} & \textbf{0.669} & \textbf{0.703} &  \textbf{0.655} & 0.626 & 0.630 &0.628  \\
      \hline
      \small Cartoon   & (0.32) \thumb{\cartoonimg} & 0.639 & \textbf{0.683} &  0.593& 0.618 &  0.647 & 0.507 \\
      \hline
      \small Sketch   & (0.14) \thumb{\sketchimg} & \textbf{0.625 } & \textbf{0.606} & 0.414 & 0.554  &  0.581 & 0.337  \\
      \hline
      %
      \hline
      \multicolumn{7}{|l|}{\textbf{ratio:} 20 images per class} \\
      \hline
      \small Art   & (0.53) \thumb{\artimg} & \textbf{0.870} & \textbf{0.851}  &  \textbf{0.861} & 0.729 &  0.849 & 0.724 \\
      \hline
      \small Cartoon   & (0.32) \thumb{\cartoonimg} & \textbf{0.912} & \textbf{0.893} & 0.841 & 0.820 &  0.887 & 0.709 \\
      \hline
      \small Sketch   & (0.14) \thumb{\sketchimg} & \textbf{0.852} &  0.638&  0.638 & 0.766 &  0.801 & 0.542\\
      \hline
      %
      \hline
      \multicolumn{7}{|l|}{\textbf{ratio:} 200 images per class}  \\
      \hline
      \small Art   & (0.53) \thumb{\artimg} & \textbf{0.906} & \textbf{0.906} &  0.823 &  0.791&  0.887 & 0.746 \\
      \hline
      \small Cartoon   & (0.32) \thumb{\cartoonimg} &  \textbf{0.958} &  0.956&  0.952  & 0.868 & 0.956 & 0.925\\
      \hline
      \small Sketch   & (0.14) \thumb{\sketchimg} & \textbf{0.924} &   \textbf{0.924 } & 0.890 & 0.767 &  0.916 & 0.875 \\
      \hline
    \end{tabular}
  }
  \vspace{0.05cm}
  \caption{\label{tab:detailres}Break-down of results comparing our proposed \ftflex{}, \ftflexfast{} and \ftflexfaster{}, to fine-tuning baselines, \ftall{} and \ftfc{}.
    In each table, the first column lists each \dshift{source}{target} domain shifts, with  the base accuracy reached by the pretrained source network on the target test set.
    Bold entries indicate the score is better than that of \textit{all }baselines (\texttt{ft-}). 
    For space reason, we only report results for a specific subsampling ratio for settings other than PACS (roughly 30 images per class for MNIST, 20 for CIFAR, 12 for ILSVRC). Full results are  in the supplemental material.}
  \end{minipage}
  ~
  \begin{minipage}[c]{0.5\textwidth}
  \resizebox{\linewidth}{!}{%
    \begin{tabular}{|l@{\hspace{ \firstcolstep}}c||c|c|c||c|c|c|c|}
      \hline
      \multicolumn{2}{|c||}{\multirow{2}{*}{\small \textbf{MNIST \thumb{\mnistimg}}}} & \multicolumn{3}{c||}{\small \ftflex{}} & \multicolumn{4}{c|}{\small \texttt{ft-}} \\
      \cline{3-9}
                                                                                      & &  \footnotesize \texttt{flex} & \footnotesize \texttt{fast} & \footnotesize \texttt{faster} & \footnotesize \texttt{fc (1)} & \footnotesize \texttt{fc (2)} & \footnotesize \texttt{ss}& \footnotesize \texttt{all} \\
      \hline
      \small Blurry  & (0.75) \thumb{\blurrymnistimg} & 0.926 & 0.926 & 0.926 & 0.921 & \textbf{0.928} &  \textbf{0.928} & 0.921 \\
      \hline
      \small Occluded  & (0.58) \thumb{\occludedmnistimg} & \textbf{0.806} & \textbf{0.806} & 0.801 & 0.785 & 0.801 &  0.792 & \textbf{0.806} \\
      \hline
      \small MNIST-M  & (0.44) \thumb{\mnistmimg} & \textbf{0.683} & \textbf{0.683} & 0.671 & 0.615 & 0.670 &  0.675 & \textbf{0.683} \\
      \hline
      \small SVHN  & (0.21) \thumb{\svhnimg} & \textbf{0.669} & \textbf{0.669} & 0.572 & 0.451 & 0.595 &  0.657 & \textbf{0.669} \\
      \hline
      \small Transf. \footnotesize (rnd)   & (0.32) \thumb{\rndtransformmnistimg} & \textbf{0.644} & \textbf{0.644} & \textbf{0.644} & 0.573 & 0.638 &  0.624 & 0.625 \\
      \hline
      \small Transf. \footnotesize (fix)  & (0.16) \thumb{\fixtransformmnistimg} & \textbf{0.908} & 0.887 & 0.879 & 0.839 & 0.875 &  0.866 & 0.887 \\
      \hline
    \end{tabular}
  }
  \vspace{0.1cm}

  \resizebox{\linewidth}{!}{%
    \begin{tabular}{|l@{\hspace{ \firstcolstep}}c||c|c|c||c|c|c|c|}
      \hline
      \multicolumn{2}{|c||}{\multirow{2}{*}{\small \textbf{CIFAR \thumb{\cifarsrcimg}}}} & \multicolumn{3}{c||}{\small \ftflex{}} & \multicolumn{4}{c|}{\small \texttt{ft-}} \\
      \cline{3-9}
                                                                                         & &  \footnotesize \texttt{flex} & \footnotesize \texttt{fast} & \footnotesize \texttt{faster} & \footnotesize \texttt{fc (1)} & \footnotesize \texttt{fc (2)} & \footnotesize \texttt{ss} &  \footnotesize \texttt{all} \\
      \hline
      \small Blurry  & (0.32) \thumb{\blurrycifarimg} & \textbf{0.577} & \textbf{0.577} & 0.512 & 0.444 & 0.501 &  0.569 & \textbf{0.577} \\
      \hline
      \small Noisy  & (0.54) \thumb{\noisycifarimg} & \textbf{0.624} & \textbf{0.624} & \textbf{0.624} & 0.583 & 0.597 &  0.618 & 0.621 \\
      \hline
      \small QuickDraw   & (0.29) \thumb{\quickdrawimg} & 0.518 & 0.517 & 0.517 & 0.475 & \textbf{0.525} & 0.495 & 0.501 \\
      \hline
    \end{tabular}
  }
  \vspace{0.1cm}
  
  \resizebox{\linewidth}{!}{%
    \begin{tabular}{|l@{\hspace{ \firstcolstep}}c||c|c|c||c|c|c|c|}
      \hline
      \multicolumn{2}{|c||}{\multirow{2}{*}{\small \textbf{QuickDraw  \thumb{\quickdrawsrcimg}}}} & \multicolumn{3}{c||}{\small \ftflex{}} & \multicolumn{4}{c|}{\small \texttt{ft-}} \\
      \cline{3-9}
                                                                                                  & &  \footnotesize \texttt{flex} & \footnotesize \texttt{fast} & \footnotesize \texttt{faster} & \footnotesize \texttt{fc (1)} & \footnotesize \texttt{fc (2)} & \footnotesize \texttt{ss} & \footnotesize \texttt{all} \\
      \hline
      \small Blurry  & (0.19) \thumb{\blurryquickdrawimg} & 0.642 & 0.631 & 0.560 & 0.426 & 0.468 &  \textbf{0.707} & 0.631 \\
      \hline
      \small Noisy   & (0.63) \thumb{\noisyquickdrawimg} & 0.801 & 0.801 & 0.795 & 0.788 & 0.792 &  \textbf{0.805} & 0.801 \\
      \hline
      \small CIFAR  & (0.20) \thumb{\cifarimg} & \textbf{0.424} & \textbf{0.424} & 0.401 & 0.333 & 0.347 &  0.388 & \textbf{0.424} \\
      \hline
    \end{tabular}
  }
  \vspace{0.1cm}

  \renewcommand{\arraystretch}{1.05}
  \resizebox{\linewidth}{!}{%
    \begin{tabular}{|l@{\hspace{ \firstcolstep}}c||c|c|c||c|c|c|}
      \hline
      \multicolumn{2}{|c||}{\multirow{2}{*}{\small \textbf{ILSVRC \thumb{\ilsvrcimg}}}} & \multicolumn{3}{c||}{\footnotesize \ftflex{}} & \multicolumn{3}{c|}{\footnotesize \texttt{ft-}} \\
      \cline{3-8}
                                                                                        & &  \footnotesize \texttt{flex} & \footnotesize \texttt{fast} & \footnotesize \texttt{faster} & \footnotesize \texttt{fc} & \footnotesize \texttt{ss} &  \footnotesize \texttt{all} \\
      \hline
      \small YUV   & (0.84) \thumb{\yuvimg} & \textbf{0.893} & \textbf{0.893} & \textbf{0.893} & 0.835 &  0.699 & 0.808 \\
      \hline
      \small HSV  & (0.38) \thumb{\hsvimg} & \textbf{0.856} & \textbf{0.856} & \textbf{0.856} & 0.533 &  0.646 & 0.687 \\
      \hline
      \small  Scaling \footnotesize(stretch)  & (0.44) \thumb{\fixedscaleimg} & \textbf{0.724} & \textbf{0.696} & \textbf{0.696} & 0.502 &  0.584 & 0.653 \\
      \hline
      \small  Scaling \footnotesize(sym.)  & (0.52) \thumb{\fixedkaleidoimg} & \textbf{0.770} & \textbf{0.757 }& \textbf{0.757 }& 0.663 &  0.650 & 0.716 \\
      \hline
      \small  Rotation  & (0.74) \thumb{\fixedrotationimg} & \textbf{0.826} & \textbf{0.832} & \textbf{0.812} & 0.667 & 0.652 & 0.771 \\
      \hline
    \end{tabular}
  }
  \end{minipage}
%
%
%
\end{table*}

\begin{figure*}[tb]
  \centering \includegraphics[width=0.92\textwidth, height=4.75cm]{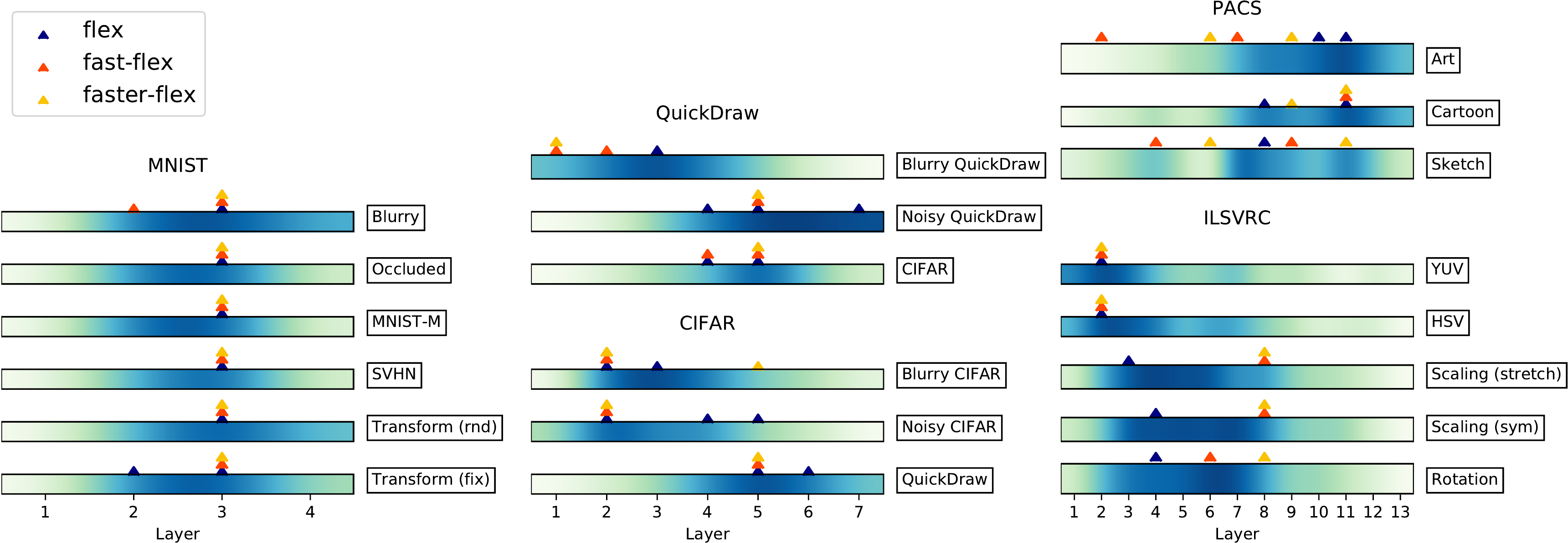}
  \caption{\label{fig:layerpick} Individual units selected by \ftflex{}, \ftflexfast{} and \ftflexfaster{}, based on \textit{validation} set accuracy. Triangles denote actual picks for \ftflex{}, \ftflexfast{} (if ignoring the option to fine-tune all units) and \ftflexfaster{}. The background values is obtained by summing the selection ranks of each unit across ratios, based on their\textit{ test performance}: in other words, the darker the color, the best performance fine-tuning this unit yields on the test set.
    We observe that \method{}'s  selection criterion generally chooses the best performing unit. The two variants' choices are more scattered, but overall positively correlate with \method{}'s decisions.
  }
\end{figure*}

In \hyperref[tab:detailres]{Table \ref{tab:detailres}} we compare the proposed method and baselines on the MNIST, CIFAR and ILSVRC-based settings, for one subsampling ratio of the target training set.
Results for other ratios show similar trends and are  available in the supplemental material. 
For the more challenging PACS scenario, which exhibits both a strong visual shift and slight semantic labels shift from the source task,  we report complete results across all subsampling ratios. 
We observe that \ftflex{} outperforms fine-tuning baselines in almost all settings. It very rarely loses to the \ftfc{} baseline, but is sometimes tied with \ftall{}, which is a subcase of \ftflex{} and \ftflexfast{} through the selection criterion.
More precisely, over all subsampling ratios and domain shifts we have in total 72 transfer scenarios. Out of these, the two overall best methods are \ftflex{} and \ftflexfast{}, achieving best accuracy 60 and 41 times respectively.
Compared to this, \ftall{} only reaches the best accuracy 26 times, mostly for large sample size and medium-sized networks. It consistently loses due to overfitting in other scenarios.
More interestingly, in terms of absolute values, we observe that when \ftflex{} strictly wins, \ie when it reaches the best accuracy and not in a tie with \ftall{}, it typically does so by a higher margin than  in the reverse scenario, \ie  when one of the baselines strictly wins.  
We detail our main observations in the rest of the section.

\medskip\noindent\textbf{Comparison to baselines.}
In the medium network or large sample size settings, \method{} expectedly generally chooses to fine-tune all layers, \ie \ftflex{} recovers \ftall{}. 
However, as the dataset size to network depth ratio decreases, fine-tuning all layers becomes more prone overfitting. 
In that case, \ftflex{} prefers to fine-tune a specific unit, which generally performs better than the \ftfc{} baseline.
More generally, the behavior of \ftfc{} strongly correlates  with the difficulty of the input domain shift:
it performs best in settings where the source domain early layers generalize well to the target domain, \eg in the noisy CIFAR  setting where the small additive random noise does not impact activations significantly.
When the domain shift is more pronounced however, \ftfc{} is often outperformed by \ftflex{}, \ftflexfast{} and \ftflexfaster{} which pick a more adequate unit to tune. 
This shows there is a benefit to having the method pick the best unit to fine-tune, rather than restricting transfer learning to the last fully-connected layers.
%
These conclusions also hold for \ftss{}, although it provides a much stronger baseline than \ftfc{} and is sometimes on-par or outperforms the faster flextuning variants. However, its performance seems to  depend on the type of domain shift: For instance, \ftss{} performs moderately well  on the colorized-ILSVRC setting. We attribute it to the fact that this setting involves a recombination of the channels which is not well captured by affine transformations of the parameters. Finally, \ftflex{} and its variants are  easier to implement in practice as they do not introduce additional parameters nor require to know how  layers actually operate.

%
%
%
%
%
%
%
%
%

\medskip\noindent\textbf{Selecting the best unit.}
We observe that the most promising unit selected by \method{} is often an intermediate one and does not follow an obvious pattern, showing that different domain shifts  affect layer representations at different depths of the architecture: This is illustrated in \hyperref[fig:layerpick]{Figure \ref{fig:layerpick}}.
On the same figure, we see that \methodsnd{} and \methodtrd{} are good approximations of \method{} as they often pick similar units. 
Similarly in \hyperref[tab:detailres]{Table \ref{tab:detailres}}  we observe  that they both often outperform fine-tuning baselines, although still being somewhat subpar to \ftflex{}.
This shows that  only a few gradient updates, as is done in \methodtrd{}, are enough to pin-point relevant units.
%

%
\medskip\noindent\textbf{Effect of the domain shift on \ftflex{}.}
From \hyperref[fig:layerpick]{Figure \ref{fig:layerpick}}, we distinguish three input domain shifts categories: 
For local \textit{pixel-level transformations}, such as noisy CIFAR, or YUV/HSV ILSVRC, \method{} tends to choose early units. 
This coincides with the fact that (i) early layers are most affected by local pixel-level changes, and (ii) such transformations are easy to correct in early layers: \eg YUV is  a linear transformation of RGB.
For \textit{geometric affine transformations}, \method{}  picks more central units of the architecture. 
In fact, such transformations do not change the global appearance of images and, moreover, most modern deep learning architecture are trained for invariance to small geometric manipulations  (\eg flip, rotations) via synthetic data augmentation, hence earlier layers are more easily transferable across these  domain shifts.
The free-transform  scenarios are harder to generalize: First, we observe that natural images features transfer particularly well across various domains. As such, \method{} often picks later layer in the architecture for general transforms scenarios with natural images as their source domain, \eg \dshift{photo}{\{art, cartoon, sketch\}}.
However, this does not seem to be the case in the reverse scenario, \eg \dshift{QuickDraw}{CIFAR} and \dshift{MNIST}{SVHN}, which indicates that features learned from the simple structure and particular distribution of binary sketches do not generalize as well to natural images.
Second, in some complex settings such as PACS, it can be the case that\textit{ two non-consecutive units} are good fine-tuning candidates. 
This suggests that units sometimes interact in complex patterns and that 
considering combination of units rather than single ones 
is an interesting future direction.

\subsection{Retrieval Experiments}
A  benefit of fine-tuning the last layer only is that it preserves a common feature representation across domains. 
However this property breaks in our setting: Images visually different from the training set fall out of the usual operation zone of the feature extractor. One can still learn a good classifier from these features \cite{kitchensinks}, but the representations themselves are meaningless with respect to the initial source domain. 
On the other hand, tuning an intermediate unit instead could help to ``mend'' the representation. To evaluate this, we use a retrieval experiment: We extract features for the initial source validation domain through the source network, and for the target domain through the flextuned or finetuned network. 
For each target sample, we  retrieve its top-$k$ nearest neighbors in the source domain and consider them correctly retrieved if they share the same semantic class, and evaluate the average precision (AP@$k$).
For space reasons, we only report AP@10, on the most challenging scenarios, in \hyperref[tab:retrieval]{Table \ref{tab:retrieval}}. Full results are reported in the supplemental material. 
The results follow our previous observations: For small networks  (MNIST, CIFAR) the result of fine-tuning all layers is often better aligned with the initial representations. However for the larger architectures, tuning an intermediate unit better recover the initial source embedding space as shown in \hyperref[tab:retrieval]{Table \ref{tab:retrieval}}.

\begin{table}
  \centering 
  \resizebox{\linewidth}{!}{%
    \begin{tabular}{|c||c|c|c|c|c|}
      \hline
      & YUV  \thumb{\yuvimg} & HSV  \thumb{\hsvimg} & Art     \thumb{\artimg} &  Cartoon \thumb{\cartoonimg} & Sketch  \thumb{\sketchimg}\\
      \hline
      \ftfc & 0.80 & 0.47 & 0.66 & 0.53 & 0.39\\
      \hline
      \ftall & 0.75 & 0.52 & 0.80 & 0.83 & 0.85\\
      \hline
      \ftflex & 0.85 & 0.85 & 0.88 & 0.86 & 0.86\\
      \hline
    \end{tabular}
  }
  \caption{mAP@10 retrieval results for the  fine-- and flex-tuned network embeddings, queried against the source domain embeddings}\label{tab:retrieval}
\end{table}


\subsection{Towards pixel-level adaptation}
%
%
An alternative to tuning a  pre-trained network is to instead learn to map target samples back to the source domain while keeping the network's weights untouched; This has the advantage of only depending on the domain shift and not on the architecture.
Such \textit{image-to-image mapping} modules have been studied 
 for domain adapation, but typically require data from 
 both source and target domains~\cite{45924, CycleGAN2017}.

\begin{figure}[t]
\begin{minipage}[c]{0.3\linewidth}
\centering
        \includegraphics[width=\linewidth]{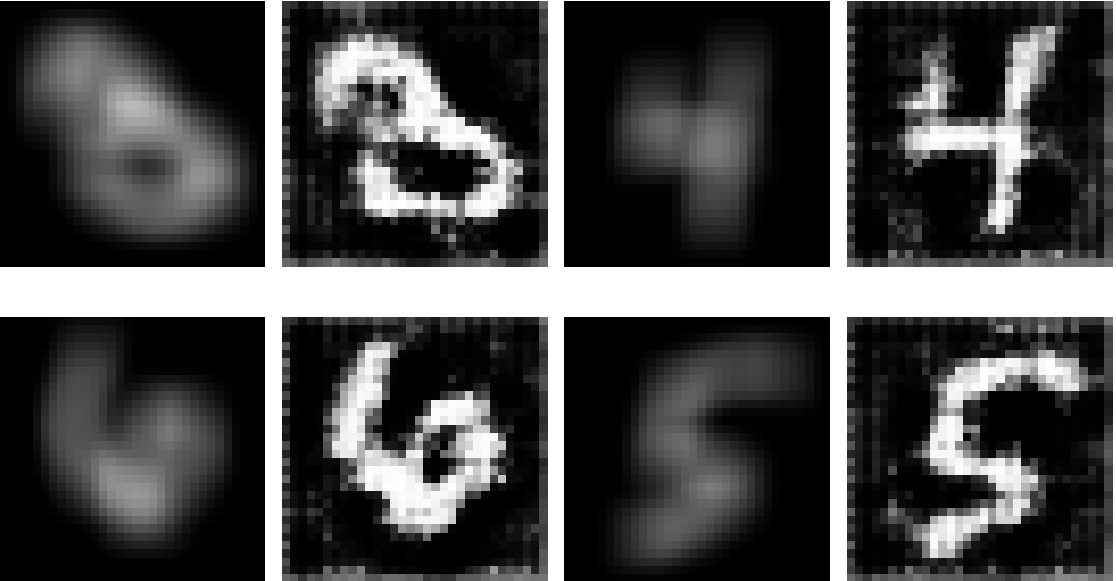}

\footnotesize (a) Blurry MNIST
  \end{minipage}
  ~
\begin{minipage}[c]{0.3\linewidth}
\centering
        \includegraphics[width=\linewidth]{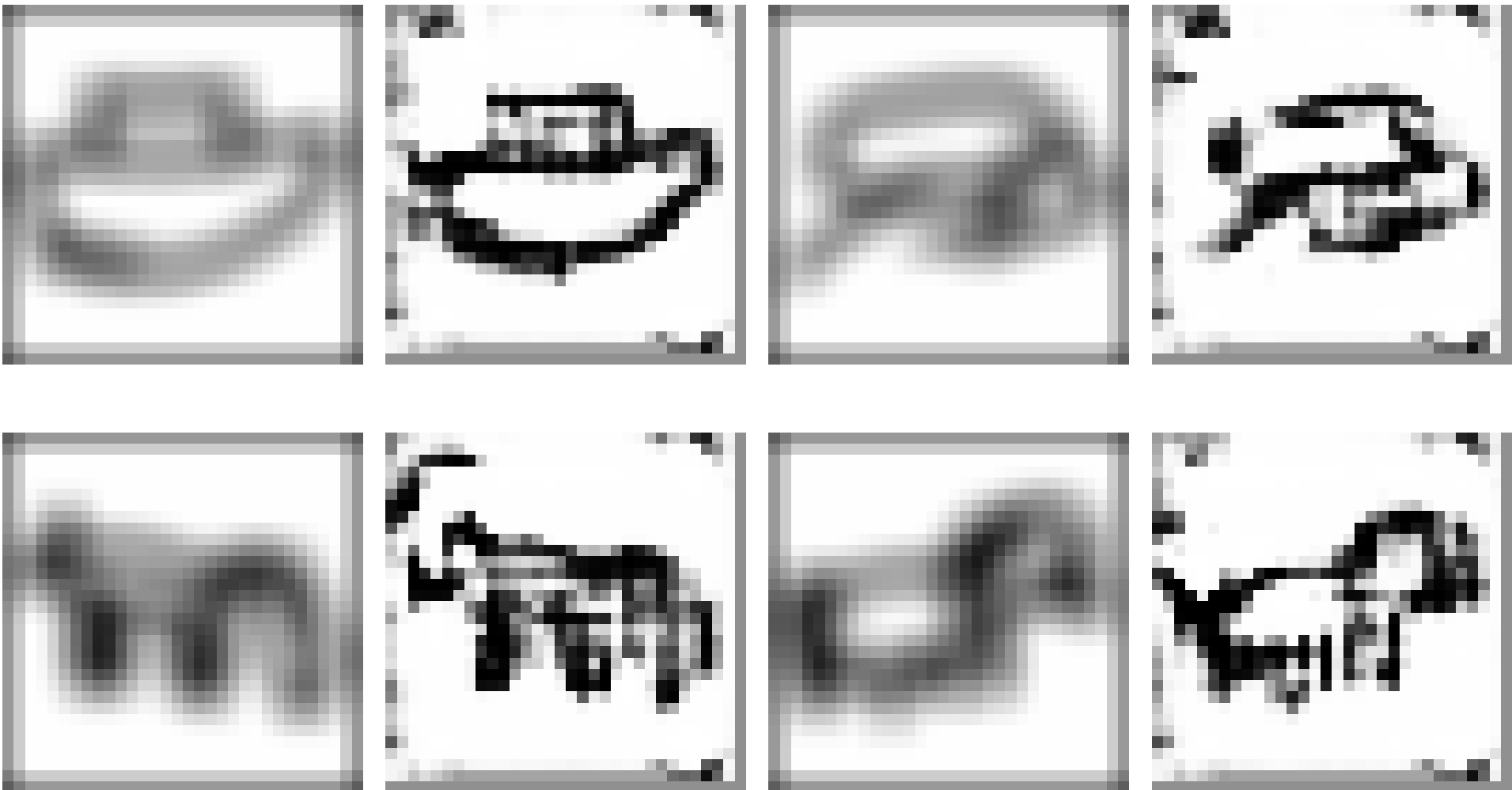}

\footnotesize (b) Blurry QuickDraw
  \end{minipage}
  ~
\begin{minipage}[c]{0.3\linewidth}
\centering
        \includegraphics[width=\linewidth]{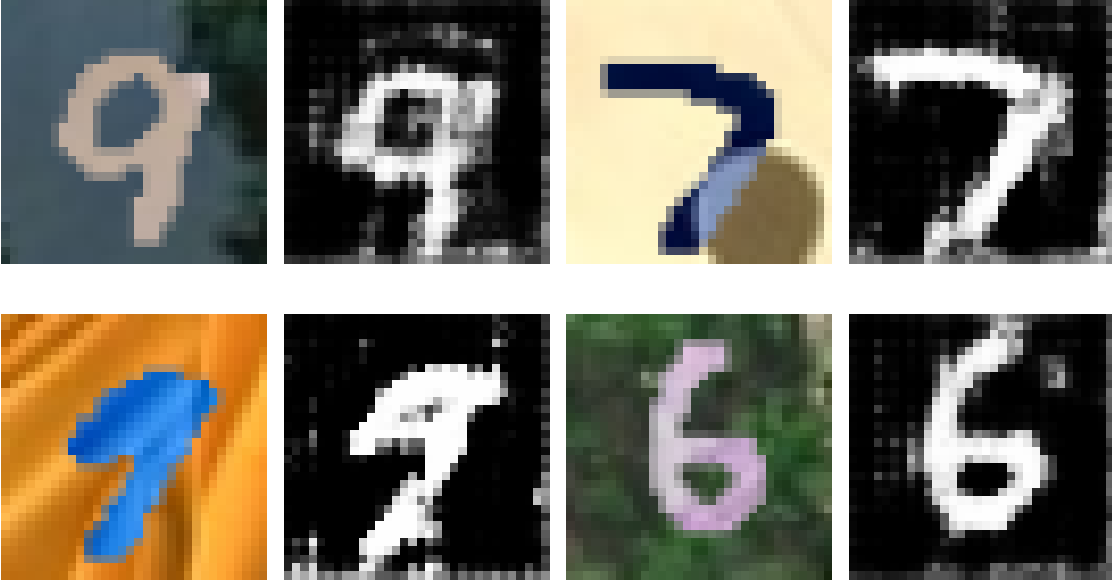}

\footnotesize (c) MNIST-M
  \end{minipage}

\begin{minipage}[c]{0.3\linewidth}
\centering
        \includegraphics[width=\linewidth]{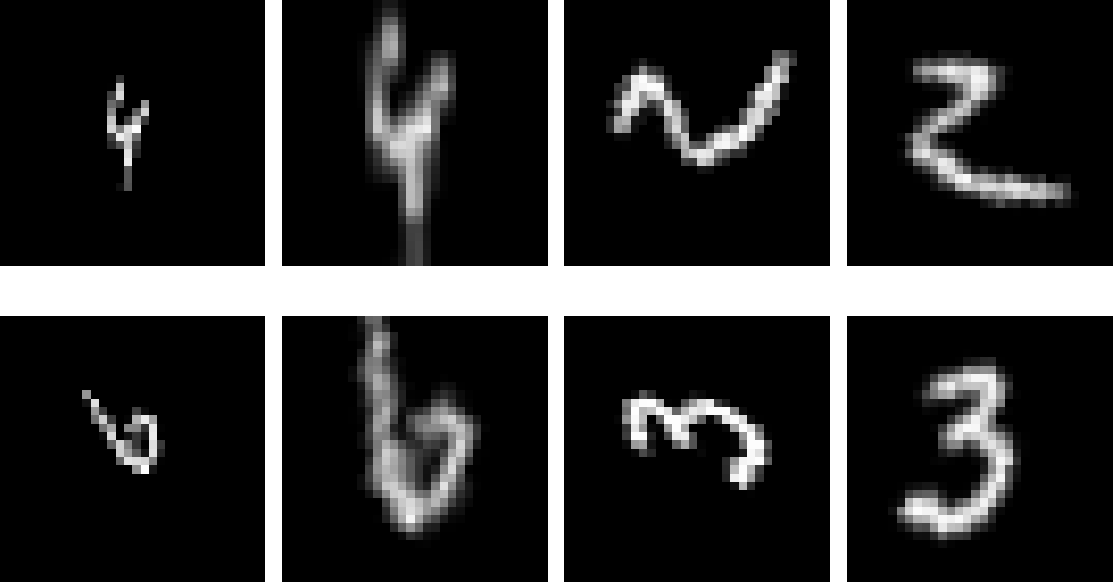}

\footnotesize (d) Transformed (rnd)
  \end{minipage}
  ~
\begin{minipage}[c]{0.3\linewidth}
\centering       
        \includegraphics[width=\linewidth]{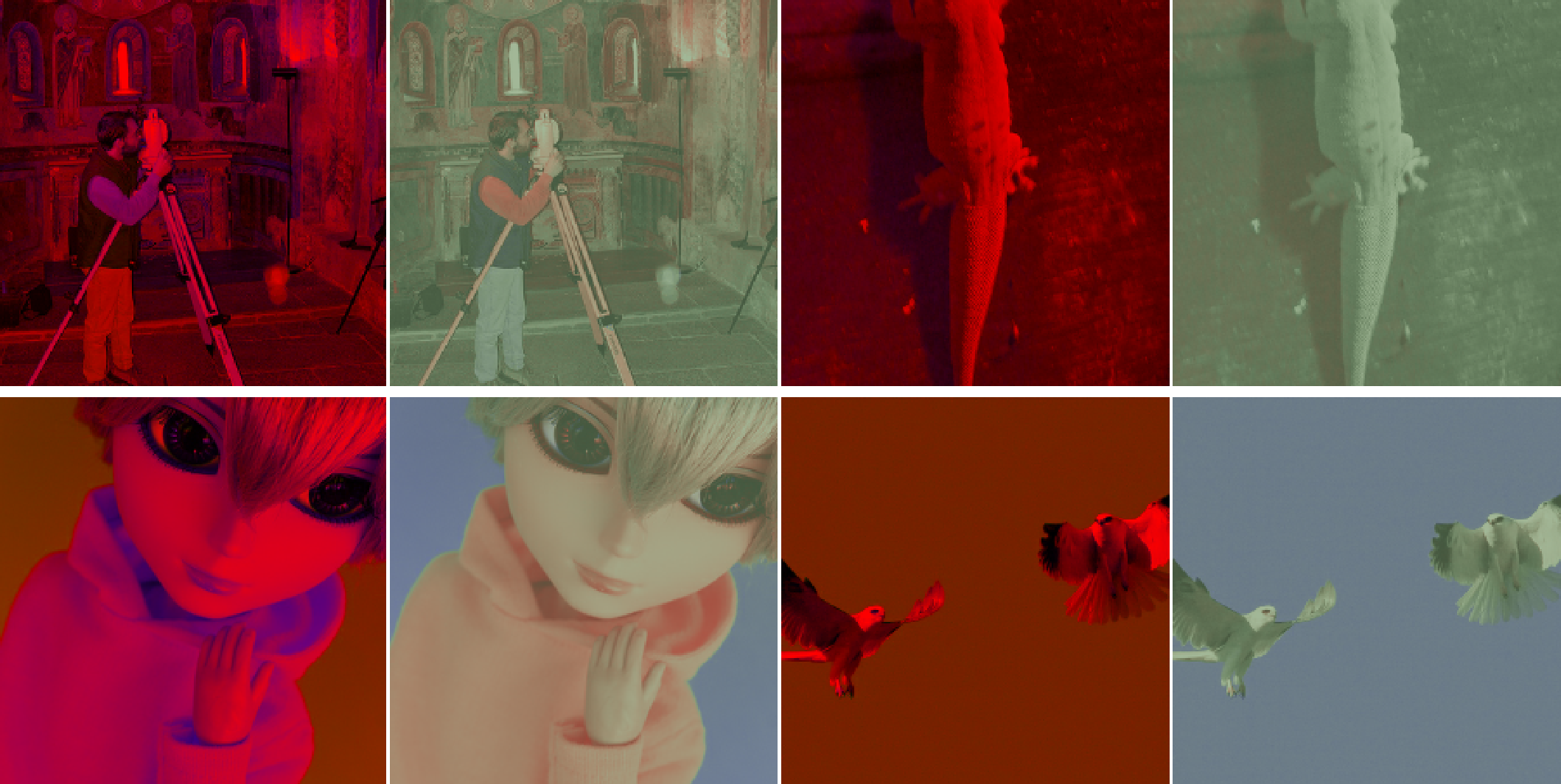}

\footnotesize (e) YUV  (1 layer)
  \end{minipage}
  ~
%
%
%
\begin{minipage}[c]{0.3\linewidth}
\centering
       \includegraphics[width=\linewidth]{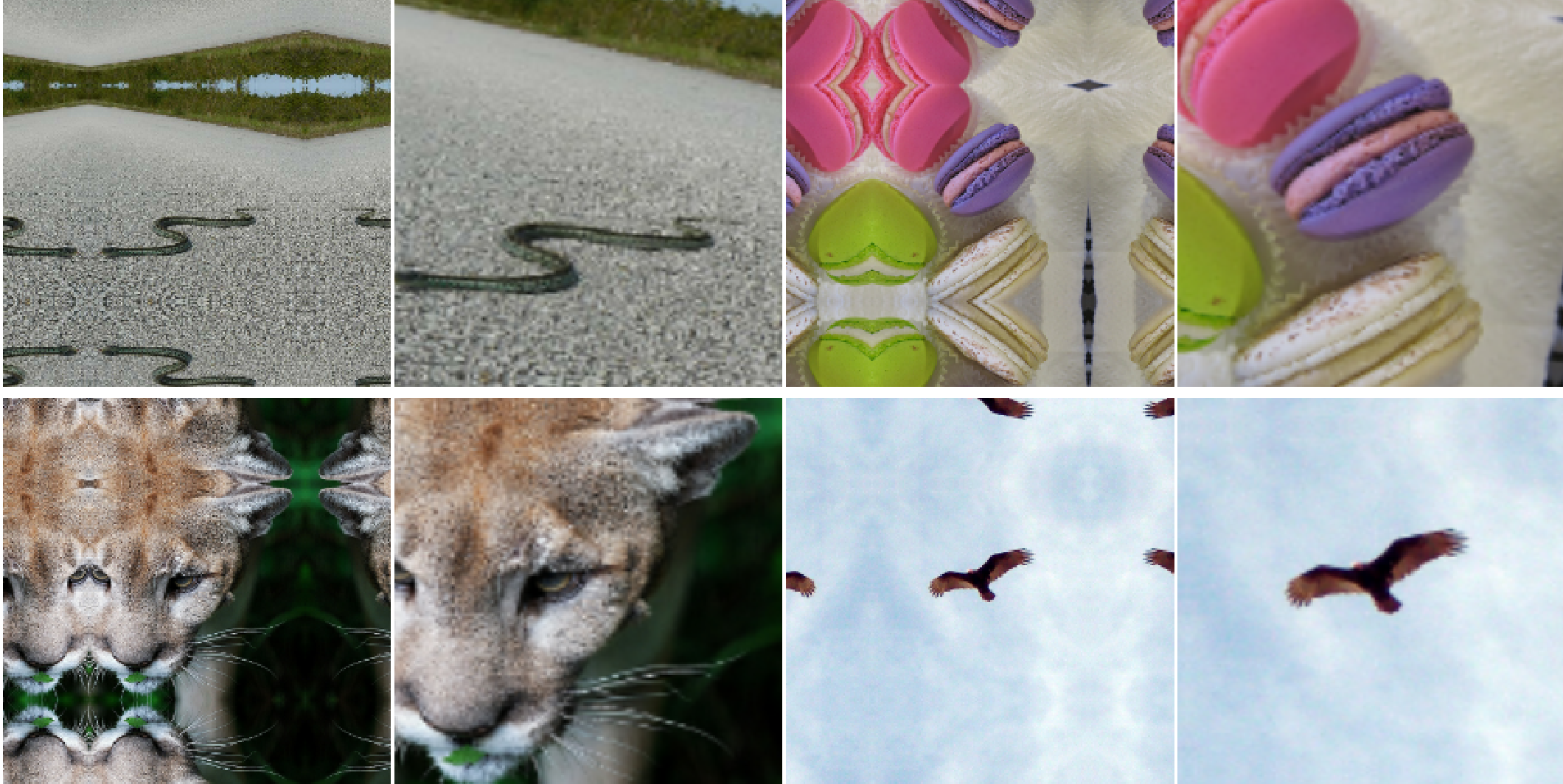}

\footnotesize (f) Scaling {\tiny (symmetric)}
  \end{minipage}
%
%




  \smallskip
\caption{Example images generated by the pre-processing module. For each pair, the left image is the input from the target domain and the right one is the pre-processed output.
%
%
}\label{fig:gninut}
\end{figure}

 Building on this idea, we introduce an \emph{image-to-image transformation unit} as
 a pre-processing module  before the feature  extraction phase of the pre-trained source network.
 The resulting architecture is considered as a new model selection option for \method{}, where only the
 image-to-image unit's weights are trained and the rest of the network is frozen.
%
 We implement this image-to-image unit  as a small Pix2Pix network~\cite{pix2pix} except in a few secnarios where we leverage our prior knowledge of the domain shift: 
For example, color channel transformations occur pixel-wise,
thus  we build the preprocessing module for YUV and HSV ILSVRC with 1x1 convolutions. 
Similarly, for geometric transforms,
we use a Spatial Transformer Network~\cite{stn}.
\hyperref[fig:gninut]{Figure \ref{fig:gninut}} shows exemplary 
outputs of the learned image-to-image units. Quantitative results are  in the supplemental material.
The specialized pre-processing modules performs  
very well for simple parametric transformations, 
%
and results are also encouraging  
on simple domain shifts  such as \emph{blur}, \emph{noise} and 
added \emph{random background}. 
We believe this to be a consequence of the skip connections 
in the Pix2Pix architecture, which enforce local pixel constraints between the input and output.
In all these succesful cases, \method's selection criterion also selects  
the image-to-image unit as the most promising unit to tune. 
%
%
For  complex transformations, \eg \dshift{photo}{sketch}, 
the pre-processing module performs poorly. 
Nevertheless,  \method is able to notice 
this and falls back to one of the other units to adapt.
%

\section{Conclusions}
\label{sec:conclusions}
We introduce a new transfer learning method for neural networks, 
\emph{\method}, that adapts a pre-trained network to a new domain by 
 tuning just a single network unit (\eg a layer or 
block  layers).
Our experiments on a variety of scenarios show that this is a surprisingly 
strong adaptation technique, as long as the right unit 
is chosen. 
Specifically, we study the case where output classes stay 
consistent but the input data characteristics change, 
potentially dramatically, \eg from  images to 
sketch drawings.
We find that, contrary to common practice, it is then rarely the last 
fully-connected unit, but rather an intermediate 
or early unit, that leads to the best adaptation results,
and \emph{\method{}} reliably identifies it.
We also introduce two accelerated 
variants that 
perform almost equally good but are significantly more computationally efficient 
in selecting the unit to be fine-tuned. 


{\small
  \bibliographystyle{ieee}
  \bibliography{egbib}

\begin{thebibliography}{10}\itemsep=-1pt

\bibitem{DBLP:journals/corr/AzizpourRSMC14}
H.~Azizpour, A.~S. Razavian, J.~Sullivan, A.~Maki, and S.~Carlsson.
\newblock Factors of transferability for a generic {ConvNet} representation.
\newblock {\em IEEE Transactions on Pattern Analysis and Machine Intelligence
  (T-PAMI)}, 2016.

\bibitem{Babenko2015AggregatingLD}
A.~Babenko and V.~S. Lempitsky.
\newblock Aggregating local deep features for image retrieval.
\newblock In {\em International Conference on Computer Vision (ICCV)}, 2015.

\bibitem{phototosketch}
P.~Ballester and R.~M. Araujo.
\newblock On the performance of {GoogLeNet} and {AlexNet} applied to sketches.
\newblock In {\em Conference on Artificial Intelligence (AAAI)}, 2016.

\bibitem{45924}
K.~Bousmalis, N.~Silberman, D.~Dohan, D.~Erhan, and D.~Krishnan.
\newblock Unsupervised pixel-level domain adaptation with generative
  adversarial networks.
\newblock In {\em Conference on Computer Vision and Pattern Recognition
  (CVPR)}, 2017.

\bibitem{Cadena2018DiverseFV}
S.~A. Cadena, M.~A. Weis, L.~A. Gatys, M.~Bethge, and A.~S. Ecker.
\newblock Diverse feature visualizations reveal invariances in early layers of
  deep neural networks.
\newblock In {\em European Conference on Computer Vision (ECCV)}, 2018.

\bibitem{quickdraw}
S.~Cheema, S.~Gulwani, and J.~LaViola.
\newblock {QuickDraw}: Improving drawing experience for geometric diagrams.
\newblock In {\em Conference on Human Factors in Computing Systems (SIGCHI)},
  2012.

\bibitem{DBLP:conf/eccv/ChuMBHD16}
B.~Chu, V.~Madhavan, O.~Beijbom, J.~Hoffman, and T.~Darrell.
\newblock Best practices for fine-tuning visual classifiers to new domains.
\newblock In {\em ECCV Workshop TASK-CV: Transferring and Adapting Source
  Knowledge in Computer Vision}, 2016.

\bibitem{Donahue:2014:DDC:3044805.3044879}
J.~Donahue, Y.~Jia, O.~Vinyals, J.~Hoffman, N.~Zhang, E.~Tzeng, and T.~Darrell.
\newblock {DeCAF}: A deep convolutional activation feature for generic visual
  recognition.
\newblock In {\em International Conference on Machine Learing (ICML)}, 2014.

\bibitem{Ganin:2015:UDA:3045118.3045244}
Y.~Ganin and V.~Lempitsky.
\newblock Unsupervised domain adaptation by backpropagation.
\newblock In {\em International Conference on Machine Learing (ICML)}, 2015.

\bibitem{Ganin:2016:DTN:2946645.2946704}
Y.~Ganin, E.~Ustinova, H.~Ajakan, P.~Germain, H.~Larochelle, F.~Laviolette,
  M.~Marchand, and V.~Lempitsky.
\newblock Domain-adversarial training of neural networks.
\newblock {\em Journal of Machine Learning Research (JMLR)}, 2016.

\bibitem{gatys}
L.~A. Gatys, A.~S. Ecker, and M.~Bethge.
\newblock Image style transfer using convolutional neural networks.
\newblock In {\em Conference on Computer Vision and Pattern Recognition
  (CVPR)}, 2016.

\bibitem{gopalan2011domain}
R.~Gopalan, R.~Li, and R.~Chellappa.
\newblock Domain adaptation for object recognition: An unsupervised approach.
\newblock In {\em International Conference on Computer Vision (ICCV)}, 2011.

\bibitem{inceptionv2}
S.~Ioffe and C.~Szegedy.
\newblock Batch normalization: Accelerating deep network training by reducing
  internal covariate shift.
\newblock In {\em Journal of Machine Learning Research (JMLR)}, 2015.

\bibitem{pix2pix}
P.~Isola, J.-Y. Zhu, T.~Zhou, and A.~A. Efros.
\newblock Image-to-image translation with conditional adversarial networks.
\newblock In {\em Conference on Computer Vision and Pattern Recognition
  (CVPR)}, 2017.

\bibitem{stn}
M.~Jaderberg, K.~Simonyan, A.~Zisserman, and K.~Kavukcuoglu.
\newblock Spatial transformer networks.
\newblock In {\em Conference on Neural Information Processing Systems (NIPS)},
  2015.

\bibitem{johnson}
J.~Johnson, A.~Alahi, and L.~Fei-Fei.
\newblock Perceptual losses for real-time style transfer and super-resolution.
\newblock In {\em European Conference on Computer Vision (ECCV)}, 2016.

\bibitem{discogan}
T.~Kim, M.~Cha, H.~Kim, J.~K. Lee, and J.~Kim.
\newblock Learning to discover cross-domain relations with generative
  adversarial networks.
\newblock In {\em International Conference on Machine Learing (ICML)}, 2017.

\bibitem{adam}
D.~P. Kingma and J.~L. Ba.
\newblock Adam: {a} method for stochastic optimization.
\newblock In {\em International Conference on Learning Representations (ICLR)},
  2015.

\bibitem{imagenettransfer}
S.~Kornblith, J.~Shlens, and Q.~V. Le.
\newblock Do better {ImageNet} models transfer better?
\newblock In {\em Conference on Computer Vision and Pattern Recognition
  (CVPR)}, 2019.

\bibitem{cifar}
A.~Krizhevsky.
\newblock Learning multiple layers of features from tiny images.
\newblock Technical report, University of Toronto, 2009.

\bibitem{larsson2016learning}
G.~Larsson, M.~Maire, and G.~Shakhnarovich.
\newblock Learning representations for automatic colorization.
\newblock In {\em European Conference on Computer Vision (ECCV)}, 2016.

\bibitem{mnist}
Y.~LeCun and C.~Cortes.
\newblock {MNIST} handwritten digit database.
\newblock \url{http://yann.lecun.com/exdb/mnist/}.

\bibitem{pacs}
D.~Li, Y.~Yang, Y.-Z. Song, and T.~M. Hospedales.
\newblock Deeper, broader and artier domain generalization.
\newblock In {\em International Conference on Computer Vision (ICCV)}, 2017.

\bibitem{fcn}
J.~Long, E.~Shelhamer, and T.~Darrell.
\newblock Fully convolutional networks for semantic segmentation.
\newblock In {\em Conference on Computer Vision and Pattern Recognition
  (CVPR)}, 2015.

\bibitem{modelzoo}
{Deep Learning Model Zoo}.
\newblock \url{https://modelzoo.co/}.

\bibitem{Netzer2011ReadingDI}
Y.~Netzer, T.~Wang, A.~Coates, A.~Bissacco, B.~Wu, and A.~Y. Ng.
\newblock Reading digits in natural images with unsupervised feature learning.
\newblock In {\em NIPS Workshop on Deep Learning and Unsupervised Feature
  Learning}, 2011.

\bibitem{deepdreams}
C.~Olah, A.~Mordvintsev, and L.~Schubert.
\newblock Feature visualization.
\newblock {\em Distill}, 2017.
\newblock https://distill.pub/2017/feature-visualization.

\bibitem{private}
N.~Papernot, M.~Abadi, {\'{U}}.~Erlingsson, I.~J. Goodfellow, and K.~Talwar.
\newblock Semi-supervised knowledge transfer for deep learning from private
  training data.
\newblock In {\em International Conference on Learning Representations (ICLR)},
  2017.

\bibitem{ParKilRojSch18}
G.~Parascandolo, N.~Kilbertus, M.~Rojas-Carulla, and B.~Sch{\"o}lkopf.
\newblock Learning independent causal mechanisms.
\newblock In {\em International Conference on Machine Learing (ICML)}, 2018.

\bibitem{kitchensinks}
A.~Rahimi and B.~Recht.
\newblock Weighted sums of random kitchen sinks: Replacing minimization with
  randomization in learning.
\newblock In {\em Conference on Neural Information Processing Systems (NIPS)},
  2008.

\bibitem{yolo}
J.~Redmon, S.~K. Divvala, R.~B. Girshick, and A.~Farhadi.
\newblock You only look once: Unified, real-time object detection.
\newblock In {\em Conference on Computer Vision and Pattern Recognition
  (CVPR)}, 2016.

\bibitem{ILSVRC15}
O.~Russakovsky, J.~Deng, H.~Su, J.~Krause, S.~Satheesh, S.~Ma, Z.~Huang,
  A.~Karpathy, A.~Khosla, M.~Bernstein, A.~C. Berg, and L.~Fei-Fei.
\newblock {ImageNet Large Scale Visual Recognition Challenge}.
\newblock {\em International Journal of Computer Vision (IJCV)}, 2015.

\bibitem{saenko2010adapting}
K.~Saenko, B.~Kulis, M.~Fritz, and T.~Darrell.
\newblock Adapting visual category models to new domains.
\newblock In {\em European Conference on Computer Vision (ECCV)}, 2010.

\bibitem{shalev2014understanding}
S.~Shalev-Shwartz and S.~Ben-David.
\newblock {\em Understanding machine learning: From theory to algorithms}.
\newblock Cambridge University Press, 2014.

\bibitem{Simon2015NeuralAC}
M.~Simon and E.~Rodner.
\newblock Neural activation constellations: Unsupervised part model discovery
  with convolutional networks.
\newblock In {\em International Conference on Computer Vision (ICCV)}, 2015.

\bibitem{scaling}
Q.~Sun, Y.~Liu, T.-S. Chua, and B.~Schiele.
\newblock Meta-transfer learning for few-shot learning.
\newblock In {\em Conference on Computer Vision and Pattern Recognition
  (CVPR)}, 2019.

\bibitem{inceptionv1}
C.~Szegedy, W.~Liu, Y.~Jia, P.~Sermanet, S.~Reed, D.~Anguelov, D.~Erhan,
  V.~Vanhoucke, and A.~Rabinovich.
\newblock Going deeper with convolutions.
\newblock In {\em Conference on Computer Vision and Pattern Recognition
  (CVPR)}, 2015.

\bibitem{tensorflowhub}
{TensorFlow Hub}.
\newblock \url{https://www.tensorflow.org/hub/}.

\bibitem{tensornets}
{TensorNets}.
\newblock \url{https://github.com/taehoonlee/tensornets}.

\bibitem{Yosinski:2014:TFD:2969033.2969197}
J.~Yosinski, J.~Clune, Y.~Bengio, and H.~Lipson.
\newblock How transferable are features in deep neural networks?
\newblock In {\em Conference on Neural Information Processing Systems (NIPS)},
  2014.

\bibitem{zeilerfergus}
M.~D. Zeiler and R.~Fergus.
\newblock Visualizing and understanding convolutional networks.
\newblock In {\em European Conference on Computer Vision (ECCV)}, 2014.

\bibitem{zhang2018unreasonable}
R.~Zhang, P.~Isola, A.~A. Efros, E.~Shechtman, and O.~Wang.
\newblock The unreasonable effectiveness of deep features as a perceptual
  metric.
\newblock In {\em Conference on Computer Vision and Pattern Recognition
  (CVPR)}, 2018.

\bibitem{CycleGAN2017}
J.-Y. Zhu, T.~Park, P.~Isola, and A.~A. Efros.
\newblock Unpaired image-to-image translation using cycle-consistent
  adversarial networks.
\newblock In {\em International Conference on Computer Vision (ICCV)}, 2017.

\end{thebibliography}
}

\end{document}